\RequirePackage{fix-cm}
\documentclass[twocolumn]{svjour3}          %

\smartqed  %
\usepackage{hyperref}
\hypersetup{
    colorlinks=true,
    breaklinks=true,
    linkcolor=red,
    filecolor=magenta,      
    urlcolor=blue,
    citecolor=blue
}
\usepackage{pifont}
\usepackage{multirow}
\usepackage{amsmath}
\usepackage{rotating}
\usepackage[dvipsnames]{xcolor}
\usepackage{comment}
\usepackage{amssymb}

\usepackage{url}
\usepackage{array} %

\usepackage[T1]{fontenc}
\usepackage{tabularx}
\usepackage{mathtools}
\usepackage[caption=false]{subfig}
\usepackage{booktabs}
\usepackage{amsfonts}
\usepackage{algorithm}
\usepackage{algpseudocode}
\usepackage{graphicx}
\newcommand{\repro}{\textsuperscript{\dag}}

\newcommand{\best}[1]{\textbf{#1}}
\newcommand{\secondbest}[1]{\underline{#1}}

\begin{document}
\sloppy
\journalname{International Journal on Document Analysis and Recognition (IJDAR)}
\title{PILOT: A Promptable Interleaved Layout-aware OCR Transformer}

\author{Laziz Hamdi \and Amine Tamasna \and Pascal Boisson \and Thierry Paquet}

\institute{
L. Hamdi \and T. Paquet \at
LITIS, Rouen, Normandy, France \\
\email{laziz.hamdi@univ-rouen.fr} \\
\email{thierry.paquet@univ-rouen.fr}
\and
A. Tamasna \and P. Boisson \at
Malakoff Humanis, Paris, France \\
\email{amine.tamasna@malakoffhumanis.com} \\
\email{pascal.boisson@malakoffhumanis.com}
}

\date{Received: date / Accepted: date}

\maketitle

\begin{abstract}
Classical OCR pipelines decompose document reading into detection, segmentation, and recognition stages, which makes them sensitive to localization errors and difficult to extend to interactive querying. This work investigates whether a single compact model can jointly perform text recognition and spatial grounding on both handwritten and printed documents. We introduce \textbf{PILOT}, a 155M-parameter prompt-conditioned generative model that formulates document OCR as unified sequence generation. A lightweight depthwise-separable CNN encodes the page, and a Transformer decoder autoregressively emits a single stream of subword and quantized absolute-coordinate tokens on a 10\,px grid, enabling full-page OCR, region-conditioned reading, and query-by-string spotting within the same architecture. A three-stage curriculum, progressing from plain transcription to joint text-and-box generation and finally to prompt-controlled extraction, stabilizes training and improves spatial grounding. Experiments on IAM, RIMES~2009, SROIE~2019, and the heterogeneous MAURDOR benchmark show that PILOT achieves competitive or superior performance in text recognition and line-level detection compared with traditional OCR systems, recent end-to-end HTR models, and compact vision--language models, while remaining substantially smaller than billion-scale multimodal models. Additional evaluations on fine-grained OCR and query-by-string spotting further confirm that a unified text--layout decoder can provide accurate and efficient promptable OCR in a compact setting. To support reproducibility, we release the synthetic SROIE generator, the 500k annotated IDL/PDFA pages, the harmonized line-level annotations for IAM, RIMES~2009, and MAURDOR, and the source code at \url{https://github.com/hamdilaziz/PILOT}.
\end{abstract}


\section{Introduction}
\label{sec:intro}

Optical Character Recognition (OCR) remains a core building block of document analysis, with applications ranging from archival digitization to industrial document processing. In many practical settings, OCR is not limited to plain transcription: users may also need to localize a queried phrase, read only a selected region, or preserve spatial grounding for downstream tasks such as key--value extraction, form processing, or visual question answering. This makes document OCR not only a recognition problem, but also a text--layout grounding problem.

For clean and relatively simple layouts, classical two-stage pipelines remain highly effective. A detector first localizes text regions, and a recognizer then transcribes the extracted crops~\cite{smith2007overview}. Such systems are mature, efficient, and often sufficient when the goal is only to read pre-segmented or well-structured content. However, this decomposition also introduces an intrinsic limitation: localization and transcription are optimized separately, and errors in the first stage directly affect the second. More importantly, once one moves beyond plain page transcription to interactive use cases such as region-conditioned OCR or query-by-string spotting, the pipeline must be extended with additional components, heuristics, or post-processing stages. In that regime, the system becomes harder to train, adapt, and maintain, especially on heterogeneous handwritten and printed documents.

Recent end-to-end OCR models reduce this dependence on explicit segmentation by transcribing text directly from images~\cite{li2023trocr,coquenet2023dan}. Yet most of them produce text alone and do not preserve explicit spatial grounding. Conversely, OCR-free document understanding models and modern vision--language models can support prompt-based interaction and structured prediction~\cite{kim2022ocr,lee2023pix2struct,hu2024mplug,wei2024general}, but they are primarily designed for high-level semantic understanding rather than compact, precise OCR. In particular, recent evaluations suggest that strong multimodal reasoning ability does not automatically translate into robust handwritten text recognition or accurate localization~\cite{Liu_2024,fu2024ocrbench}. Thus, there is still a missing point in the design space: a model that remains lightweight, performs strong OCR on both handwritten and printed documents, and supports explicit spatially grounded interaction within a single architecture.

In this work, we introduce \textbf{PILOT}, a compact 155M-parameter promptable OCR model that unifies text recognition and localization as a single autoregressive generation problem. Instead of treating layout prediction as a separate detection branch, PILOT generates a mixed sequence of subword tokens and quantized coordinate tokens, allowing the same decoder to perform full-page OCR, region-conditioned reading, and query-by-string spotting. This formulation is motivated by two goals. First, it removes the need for task-specific detection heads and external coupling between localization and recognition. Second, it provides a unified interface for interactive OCR, where text content and spatial position are predicted within the same decoding process.

Our focus is therefore not to replace every strong two-stage OCR system on simple benchmarks, where such pipelines may already perform well, but to study whether a \emph{single compact model} can jointly provide transcription, localization, and promptable querying across heterogeneous document conditions. We show that this is possible with a lightweight architecture and an appropriate training curriculum.

Our contributions are threefold:
\begin{itemize}
\item We propose a lightweight layout-aware encoder--decoder that unifies full-page OCR, region-conditioned OCR, and query-by-string spotting within a single generative framework, while remaining effective on both printed and handwritten documents.
\item We systematically study design choices for mixed text--layout generation, including absolute vs.\ relative coordinate discretization, single- vs.\ dual-branch localization, grid resolution, and the ordering of text and coordinate tokens.
\item We will release a collection of 500k annotated IDL/PDFA pages, a synthetic generator for SROIE-style receipts, and harmonized line-level annotations for IAM, RIMES~2009, and MAURDOR, under their original license terms.
\end{itemize}

\begin{figure}[t]
\centering
\includegraphics[width=\linewidth]{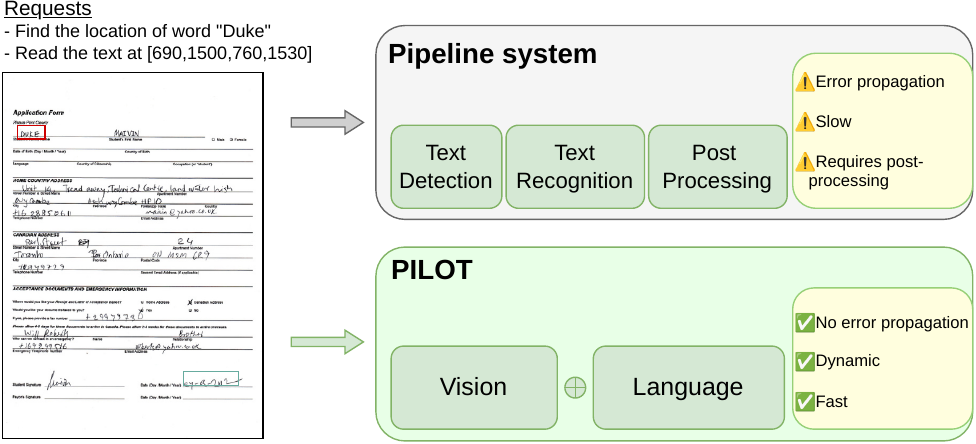}
\caption{Comparison between a conventional OCR pipeline and PILOT. Unlike multi-stage systems that separate detection, recognition, and post-processing, PILOT uses a single promptable model to jointly support text recognition and localization.}
\label{VISTA_sample}
\end{figure}

\section{Related Work}
\subsection{Pipeline‑based OCR}
\label{standard_ocr}
Early engines such as Tesseract~\cite{smith2007overview}, EasyOCR~\cite{easyocr}, PyLaia~\cite{puigcerver2018pylaia}, PERO‑OCR~\cite{kohut2021ts}, and PaddleOCR~\cite{paddleocr} process documents in two separate stages: a detector, often based on connected components or modern CNNs, predicts word boxes, and a recogniser transcribes each crop. While mature and highly optimised, these systems require domain‑specific retraining to cope with exotic fonts or handwriting and remain vulnerable to cascading errors and to false positives outside the region of interest.

\subsection{End‑to‑End, OCR‑Free Models}
\label{end-to-end_ocr}
Inspired by neural machine translation, TrOCR~\cite{li2023trocr} and DAN~\cite{coquenet2023dan} generate transcripts directly from pixels. Donut~\cite{kim2022ocr}, Dessurt~\cite{davis2022end}, Pix2Struct~\cite{lee2023pix2struct}, and DANIEL~\cite{constum2025daniel} extend this paradigm to document understanding tasks without external OCR, enabling end-to-end NER, VQA, and form understanding. Their main limitation is that they model layout only implicitly and cannot directly output precise text locations, perform region-constrained OCR, or support query-by-string extraction. 

\subsection{Large Vision--Language Models}
\label{lvlms}
\textbf{Heavyweight models} mPLUG-DocOwl~\cite{hu2024mplug} proposes a modularized training framework for multimodal LLMs that incorporates visual context. TextMonkey~\cite{liu2026textmonkey} wraps a 7.7B-parameter LLM within a 9.7B-parameter pipeline that adds visual and token-resampling modules. PaliGemma~\cite{beyer2024paligemma}, released in 3B, 10B, and 28B variants, pairs a SigLIP vision encoder with a Gemma-2B decoder~\cite{team2024gemma} and excels at image-grounded chat.\\
\textbf{Mid-size models} Florence-2-L~\cite{Xiao_florence2}, GOT-OCR 2.0~\cite{wei2024general}, FOx~\cite{liu2024focus}, and Éclair~\cite{karmanov2025eclair} offer lighter alternatives. FOx compresses a 1024$\times$1024 page into 256 image tokens and uses position-aware prompts to handle multi-page layouts, whereas Éclair jointly predicts reading order and layout markup on printed multi-page documents.\\
\textbf{Compact models} Florence-2-B~\cite{Xiao_florence2}, SmolDocling~\cite{nassar2503smoldocling}, and SmolVLM~\cite{marafioti2504smolvlm} convert PDFs into HTML or structured JSON, but they are trained primarily on synthetic or scanned \emph{printed} corpora and therefore do not target handwritten transcription.

\subsection{Coordinate-as-Token Generative Models}
\label{sec:gen-text-box}
Casting vision tasks as sequence generation began with object detection: Pix2Seq~\cite{chen2022pix2seq} represents each bounding box by four discrete coordinate tokens, and Pix2Seq~v2~\cite{chen2022pix2seqv2} extends the interface to segmentation and keypoints. In scene‑text, TESTR~\cite{zhang2022testr} predicts Bézier control points and character sequences via two Transformer decoders; UNITS~\cite{kil2023units} unifies arbitrary‑shaped boxes and introduces starting‑point prompting to exceed the trained instance count; HTS~\cite{long2024hts} further captures a four‑level hierarchy (character→word→line→paragraph).  
For documents, LayTextLLM~\cite{lu2024laytextllm} interleaves a \emph{single} learned token per box with text inside a frozen LLM, but it consumes OCR output and does not generate coordinates. 

\subsection{Text Spotting in Handwritten Documents}
Early “word-spotting’’ systems avoid full HTR/OCR by matching a query string directly to page regions. PHOCNet~\cite{PHOCNet} learns a joint embedding between word-image crops and Pyramidal Histogram-of-Characters attributes, whereas Neural Ctrl-F~\cite{ctrlfnet} formulates spotting as region-proposal + embedding and is still a strong baseline on historical manuscripts. Bi-Gram KWS~\cite{ghosh2015query} accelerates query-by-string
retrieval by indexing character bi-grams, and SegFreeKWS~\cite{retinas2023} drops explicit segmentation altogether, using character-counting and CTC re-scoring to achieve competitive mAP on IAM pages. More recently, ST-KeyS~\cite{ST-KeyS} shows that a self-supervised ViT encoder trained with masked auto-encoding can match or surpass supervised CNNs when annotation is scarce.

\begin{figure*}[t]
\centering
\includegraphics[width=\linewidth]{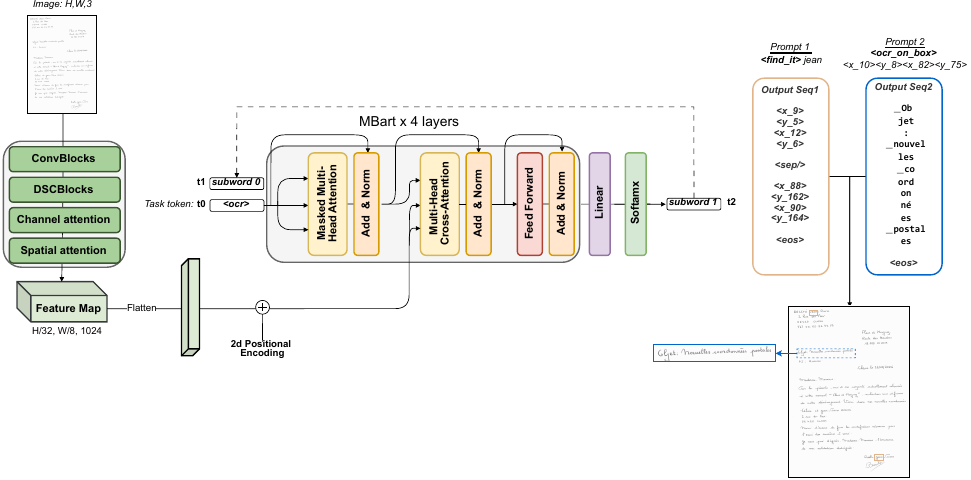}
\caption{Overview of PILOT. A depthwise‑separable CNN encodes the page, and a Transformer decoder autoregressively emits a single sequence that interleaves text and quantised bounding‑box coordinates conditioned on an optional prompt.}
\label{vista_archi}
\end{figure*}

\section{Approach}
\label{sec:approach}
\subsection{Problem Formulation}
PILOT formulates document OCR as a conditional sequence generation problem that jointly models textual content and spatial localization. Given an input page image $\mathbf{I}\in\mathbb{R}^{H\times W\times 3}$, the model predicts an ordered output sequence
\[
\mathbf{S} = (s_1,\dots,s_T),
\]
where each token $s_t$ is drawn from a unified vocabulary
\[
\mathcal{V}=\mathcal{V}_{\text{text}} \cup \mathcal{V}_{\text{box}}.
\]
The subset $\mathcal{V}_{\text{text}}$ contains subword units, whereas $\mathcal{V}_{\text{box}}$ contains discrete coordinate tokens that encode quantized page positions.

Unlike conventional OCR pipelines, which decompose the problem into text detection, line segmentation, and transcription, PILOT predicts both linguistic and geometric information within the same autoregressive stream. Each text line is serialized as a sequence interleaving coordinate and subword tokens:
\[
[x_1][y_1]\; w_1 \dots w_K\; [x_2][y_2],
\]
where $[x_1],[y_1]$ and $[x_2],[y_2]$ denote the top-left and bottom-right corners of the line bounding box, and $w_1,\dots,w_K$ are textual subword tokens. For instance, under a 10\,px quantization grid, the word \emph{boxes} associated with the bounding box $(40,60,60,200)$ is serialized as
\[
[\texttt{<x\_4>}][\texttt{<y\_6>}]\; \texttt{\_box}\; \texttt{es}\; [\texttt{<x\_6>}][\texttt{<y\_20>}],
\]
where the lexical content is represented by two subword tokens and the box coordinates are converted into discrete position tokens.

In this way, recognition and localization are learned as a single structured prediction problem.

This formulation also provides a simple prompt-based interface for switching between full-page OCR, region-conditioned transcription, and query-by-string spotting. By prepending a task-specific prompt, the same model can perform standard full-page OCR, region-conditioned transcription, or query-by-string spotting. In contrast to multi-branch approaches that recognize text and regress coordinates separately~\cite{mao2024visually}, PILOT uses a single decoder to model the dependencies between language and layout directly, which encourages coherent text--box alignment and avoids error propagation between separate modules.

\subsection{Model Architecture}
An overview of PILOT is shown in Figure~\ref{vista_archi}. The model follows a compact encoder--decoder design intended to preserve fine visual detail while remaining lightweight enough for practical OCR deployment.

The encoder is a convolutional backbone composed of Convolutional Blocks (CB) and Depthwise-Separable Convolutional Blocks (DSCB), inspired by the Vertical Attention Network (VAN) design~\cite{coquenet2022end}. This choice is motivated by two requirements that are central to document OCR. First, the encoder must retain high-resolution local evidence to resolve small characters, punctuation marks, and thin separators. Second, it must aggregate enough context to distinguish text from structured background patterns such as ruling lines, stamps, or repeated form elements. The backbone therefore combines progressive depth scaling with residual connections and SiLU activations~\cite{ElfwingSILU} to maintain stable optimization while keeping the parameter count low.

To further improve feature discrimination, we incorporate Efficient Channel Attention (ECA)~\cite{WangECAnet} and CBAM spatial attention~\cite{woo2018cbam}. In our experiments, these modules do not materially change pure text-recognition scores, but they consistently improve localization quality, which is consistent with their role in sharpening spatially informative features, especially when the model must separate foreground text from textured or cluttered backgrounds.

The encoder outputs a dense feature map that preserves fine-grained page structure. Two-dimensional sine positional encodings are added to this grid before flattening it and passing it to a four-layer Transformer decoder initialized from mBART~\cite{liu2020multilingual}. The decoder autoregressively generates the output sequence from the unified vocabulary
$\mathcal{V}=\mathcal{V}_{\text{text}} \cup \mathcal{V}_{\text{box}}$.
Textual and coordinate tokens are therefore predicted by the same projection head, allowing the decoder to reason jointly over content and location without introducing task-specific detection heads or auxiliary localization branches.

We deliberately keep the decoder shallow, with four layers, to balance capacity, speed, and robustness. In preliminary experiments, increasing decoder depth did not improve recognition or localization, while it made the model larger, slower, and more prone to overfitting. We attribute this to the moderate size of currently available OCR training corpora compared with the multi-million-sample datasets typically used to train larger generative document models such as Donut or Pix2Struct. Under this regime, a compact decoder was sufficient to model the relevant text--layout dependencies while preserving favourable inference efficiency.

\subsection{Unified Text--Layout Tokenization}
\label{subsec:tokenization}
A central component of PILOT is its tokenization scheme, which encodes textual content and spatial layout in a single discrete sequence. This design is critical because the decoder must learn not only what to transcribe, but also where each textual instance is located on the page.

Following the sequence-generation perspective introduced in object detection by Pix2Seq~\cite{chen2022pix2seq} and extended to scene text by models such as UNITS~\cite{kil2023units}, PILOT represents coordinates as tokens. However, unlike prior approaches that discretize normalized coordinates relative to the input size, we use \emph{absolute} pixel coordinates quantized on a fixed 10\,px grid. Thus, a page location is mapped to a token that represents a spatial bin in image space rather than a relative proportion of width or height.

This choice offers two practical advantages. First, it simplifies the geometry that the decoder must learn: the same token always corresponds to the same approximate page position, which facilitates consistent alignment between convolutional features and coordinate predictions. Second, it avoids repeatedly remapping the coordinate vocabulary when pages share similar physical layout but differ slightly in resolution. In our setting, the coordinate vocabulary is large enough to cover page dimensions up to approximately $5000\times5000$ pixels, which is well above the typical range encountered in our target corpora. For pages up to $1800\times2400$ pixels, the default 10\,px quantization yields around 180 distinct $x$-tokens and 240 $y$-tokens.

Each line is serialized as
\[
s_i = [x_1][y_1]\; w_1 w_2 \dots w_K\; [x_2][y_2],
\]
where $w_j\in\mathcal{V}_{\text{text}}$ are subword tokens and the coordinate tokens denote the bounding-box corners. We use distinct vocabularies for the $x$- and $y$-axes rather than a shared location vocabulary. In the ablation study, this design gives a small but consistent gain over shared coordinate tokens, while adding negligible vocabulary size relative to the text vocabulary. We also study alternative interleavings of text and coordinate tokens and find that the exact ordering has only a minor effect, whereas the separation of horizontal and vertical coordinate vocabularies is more important.

\begin{figure}[t]
\centering
\includegraphics[width=\linewidth]{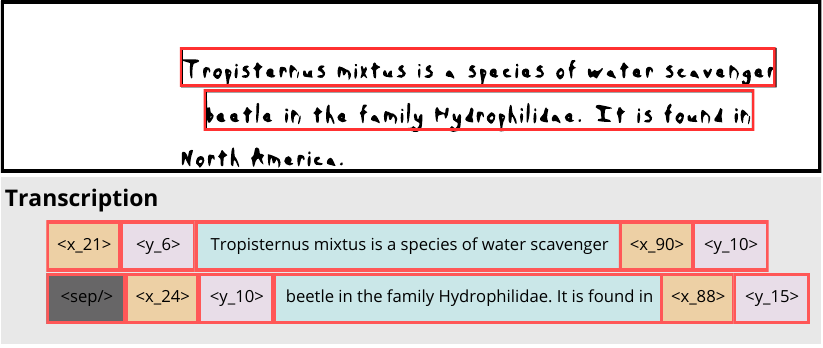}
\caption{Example of unified text--layout encoding. Each line transcription is delimited by coordinate tokens representing the corners of its bounding box.}
\label{synth_encoding_sample}
\end{figure}

A potential concern with this mixed-stream design is that interleaving text and coordinate tokens could disrupt the language-modeling behavior of the decoder. In practice, we did not observe such degradation. The staged training procedure described below first trains the model on pure text transcription, allowing the decoder to acquire a stable language prior before coordinate tokens are introduced. Empirically, adding spatial tokens in the second stage improves localization without harming recognition quality, which suggests that the model learns to separate textual and geometric token types within a shared decoding space. An example of the resulting text--layout serialization is shown in Figure~\ref{synth_encoding_sample}.

\subsection{Learning Objective and Training Curriculum}
\label{subsec:training_curriculum}
PILOT is trained end-to-end with a label-smoothed cross-entropy loss defined over the unified vocabulary:
\[
\mathcal{L} = \lambda\,\mathcal{L}_{\text{text}} + (1-\lambda)\,\mathcal{L}_{\text{box}},
\]
where $\mathcal{L}_{\text{text}}$ and $\mathcal{L}_{\text{box}}$ denote the token-level cross-entropies for textual and coordinate predictions, respectively. We set $\lambda=0.5$, which provided the best compromise between transcription accuracy and spatial precision on the validation sets. This objective directly matches the inference interface of PILOT, since both text and box tokens are generated within the same autoregressive sequence.

Optimization follows a progressive three-stage curriculum designed to stabilize training as the task complexity increases. Rather than asking the model to learn transcription, localization, and prompt control simultaneously from scratch, we introduce these capabilities in succession.

\paragraph{Stage 1: Plain transcription.}
In the first stage, PILOT is trained as a standard OCR model that predicts only textual tokens. The decoder vocabulary is restricted to $\mathcal{V}_{\text{text}}$, with no coordinate supervision. This warm-up phase is performed on a visually diverse subset of 100\,k pages drawn from the IDL and PDFA corpora. To promote appearance diversity, the subset is sampled randomly after a preliminary visual clustering step. At this stage, the encoder learns robust document features, while the decoder acquires an initial prior over text generation and reading order. Training first on pure transcription also limits interference from unstable early coordinate predictions and provides a more stable initialization for the subsequent text--layout learning stages.

\paragraph{Stage 2: Joint text--box generation.}
In the second stage, we augment the vocabulary with coordinate tokens $\mathcal{V}_{\text{box}}$ and extend the decoding length to 2048 tokens. The model is then trained with the full joint objective so that each text line is generated together with its quantized bounding box. This stage is where PILOT learns the alignment between the visual content of the page and the spatial token stream. Because the decoder already has a stable transcription prior from Stage~1, it can incorporate positional information without sacrificing recognition quality.

\paragraph{Stage 3: Prompt-controlled modes.}
In the final stage, PILOT is exposed to prompt-conditioned OCR tasks. In the \emph{region-conditioned OCR} setting, the prompt specifies a rectangular region through its corner coordinates, and the model must transcribe only the text contained inside that region. In the \emph{query-by-string} setting, the prompt provides a textual query, and the model predicts the coordinates of all matching occurrences on the page. These tasks reuse exactly the same architecture and vocabulary as full-page OCR; only the prompt prefix changes to indicate the required behavior. This preserves architectural simplicity while enabling interactive OCR capabilities.

\paragraph{Optimization details.}
All three stages are trained sequentially with AdamW and a linearly decaying learning rate. Stage~1 is optimized for approximately 200 steps with an initial learning rate of $1\times10^{-5}$. Stages~2 and~3 are then trained for 300 and 1000 epochs, respectively, with the learning rate decayed from $1\times10^{-5}$ to $1\times10^{-6}$. Each stage is initialized from the checkpoint of the previous one, ensuring a smooth transition between objectives. Training is performed on a single NVIDIA H200 GPU (140\,GB).

Because training relies on teacher forcing, the decoder is optimized under gold-prefix conditioning and is therefore not directly exposed to its own prediction errors. To reduce overfitting to this setting and improve robustness at inference time, we inject controlled noise into the supervision. With probability $0.2$, a target text token is replaced by a nearby alternative under Tree Edit Distance, exposing the decoder to mild lexical perturbations. Coordinate tokens are also jittered by $\pm 3$ quantization bins to simulate small localization errors and annotation noise. In parallel, we apply a range of image degradations to mimic common acquisition and scanning artifacts, including variations in zoom ratio and rendering DPI, perspective distortions, dilation/erosion, color jitter, Gaussian blur, Gaussian noise, sharpening, and elastic distortions. Combined with the progressive curriculum, these perturbations improve generalization and help PILOT learn a shared text--layout representation that supports both standard OCR and prompt-controlled extraction within a compact model.

\section{Experiments and Results}
\label{sec:experiments}
This section describes the experimental protocol used to train and evaluate PILOT, including the construction of the training corpus, preprocessing choices, evaluation metrics, and comparisons with existing OCR and vision--language baselines. We consider both handwritten and printed documents and evaluate the model on two complementary dimensions: \emph{text recognition} (TR) and \emph{text detection} (TD). The goal is to assess not only transcription accuracy, but also localization quality, promptability, and computational efficiency under a unified evaluation setting.

\subsection{Training Data and Preprocessing}

To train and evaluate PILOT across diverse document conditions, we constructed a large-scale corpus combining real and synthetic pages.  
All images are processed in RGB with three channels and normalized according to the mean and standard deviation of the training set of each dataset during fine-tuning.  
For pretraining and evaluation on unseen documents, we adopt the standard ImageNet mean and variance, ensuring compatibility with pretrained visual backbones.

\paragraph{Real documents}
We curated approximately \textbf{0.5\,M} pages from the Industry Documents Library (IDL) and SafeDocs PDFA corpora.  
Each file was rasterized at 200\,dpi, converted to RGB, and pages exceeding twice the A4 size were isotropically downscaled.  
Automatic deskewing was applied to correct rotations up to $\pm5^{\circ}$, and blank or low-contrast pages were discarded.  
To prevent overrepresentation of repetitive templates, pages sharing identical structure were subsampled to maintain layout diversity.  
From this pool, 1\,000 pages were manually refined to form a validation subset.

For fine-tuning and evaluation, we use the standard handwritten and printed benchmarks IAM~\cite{marti2002iam}, RIMES~\cite{grosicki2009results}, MAURDOR~\cite{brunessaux2014maurdor}, and SROIE~\cite{huang2019icdar2019}.  
As these datasets differ in granularity, we harmonized all annotations to line-level bounding boxes: IAM’s segmentation errors were corrected, we re-annotated RIMES and MAURDOR to produce consistent line boxes.  
This harmonization yields a homogeneous supervision interface across printed and handwritten domains.

\begin{figure}[t]
\centering
\subfloat[SROIE-synthetic]{%
\fbox{\includegraphics[width=0.45\linewidth]{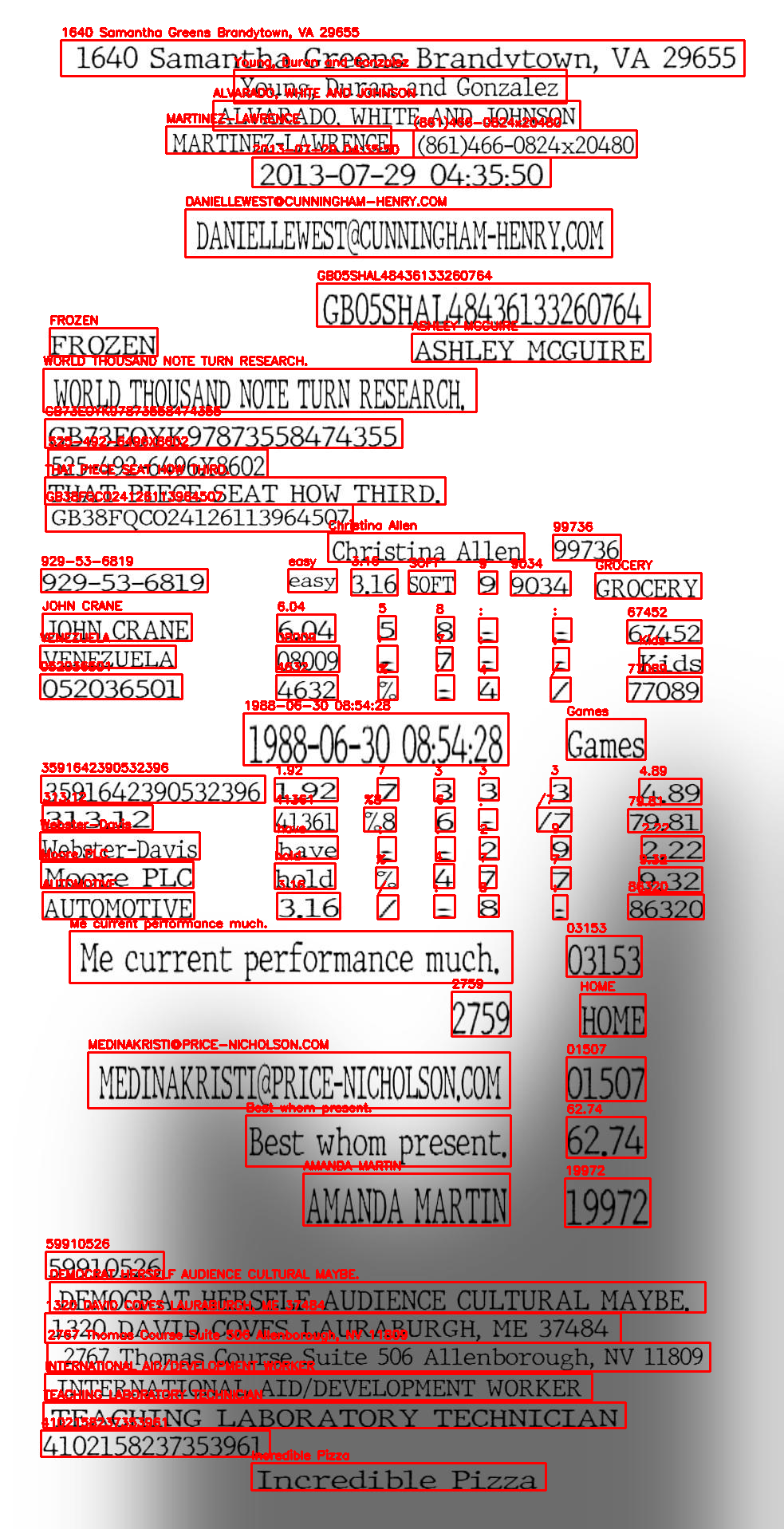}}%
\label{fig:sroie_synth}
}
\hfill
\subfloat[RIMES-synthetic]{%
\fbox{\includegraphics[width=0.45\linewidth]{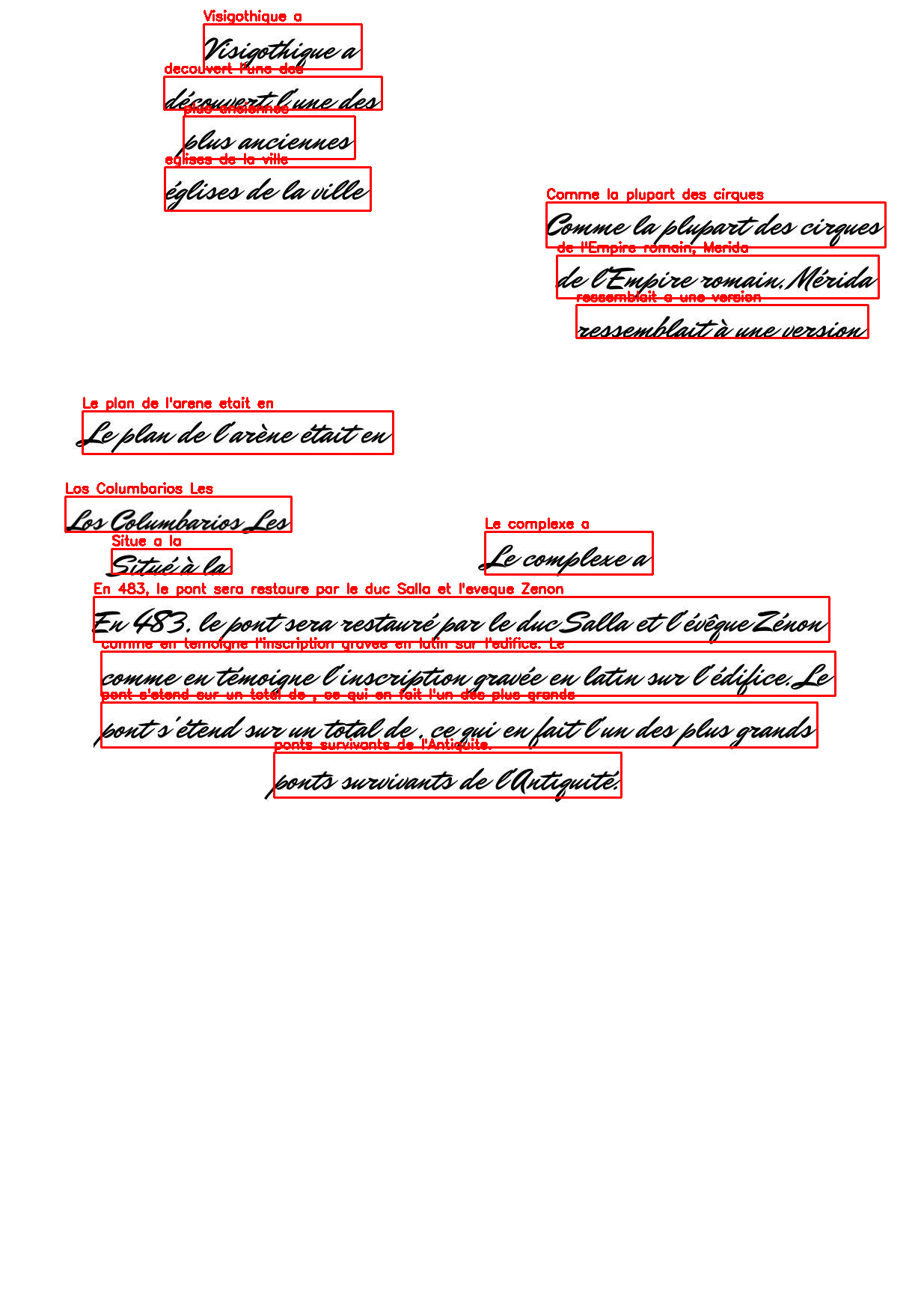}}%
\label{fig:rimes_synth}
}

\vspace{0.5em}

\subfloat[IAM-synthetic]{%
\fbox{\includegraphics[width=0.45\linewidth]{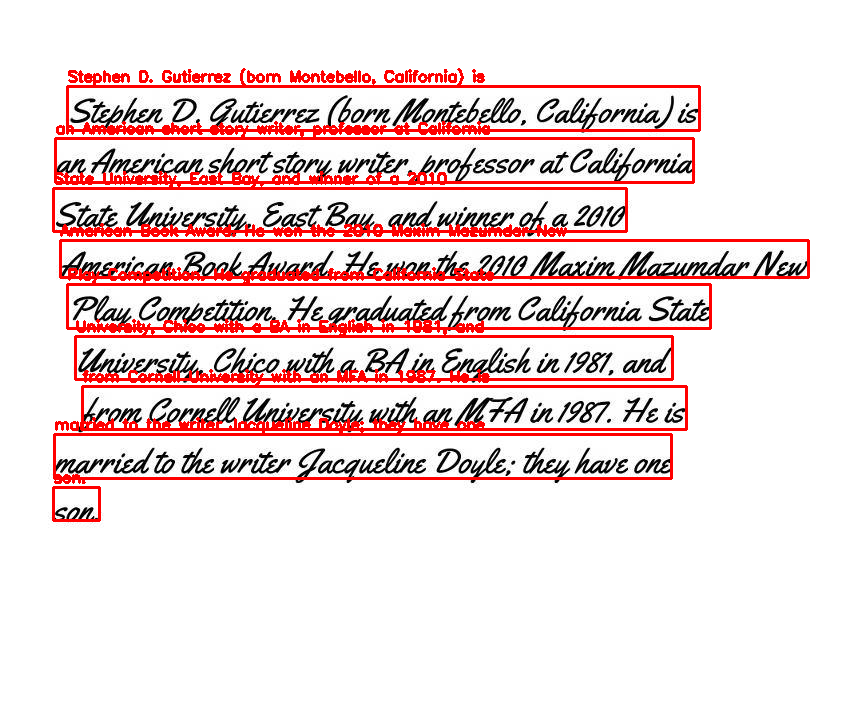}}%
\label{fig:iam_synth}
}
\hfill
\subfloat[SynthDOG]{%
\fbox{\includegraphics[width=0.45\linewidth]{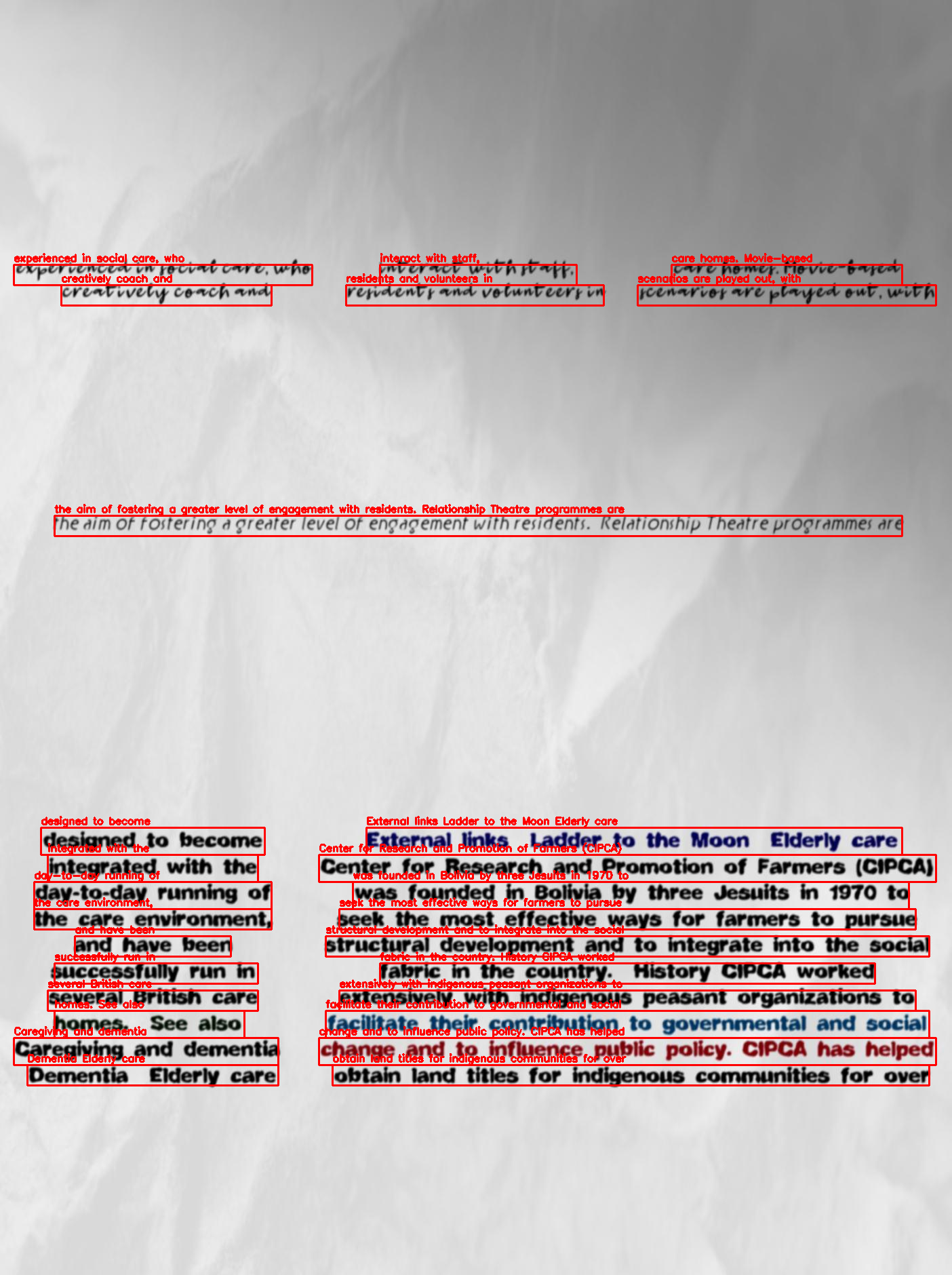}}%
\label{fig:synthdog_synth}
}

\caption{Representative synthetic document samples used for pretraining and augmentation.  
Each generator reproduces characteristic layout and visual style of its target domain while preserving precise line-level bounding boxes.}
\label{fig:synthetic_samples}
\end{figure}

\paragraph{Synthetic documents}
To complement the real data and cover a wider stylistic range, we generated large-scale synthetic corpora in English and French (see Figure~\ref{fig:synthetic_samples}).  
We employed the IAM+RIMES generator of~\cite{constum2025daniel}, extended with line-level bounding-box output, and used SynthDOG~\cite{mao2024visually} to produce an additional 100\,k bilingual pages.  
Across all synthetic sources, text rendering draws from a curated library of approximately \textbf{4\,000} fonts including serif, sans-serif, and handwriting styles, each verified to support diacritics for both languages.  
Fonts are randomly sampled per line with random perturbations in weight, size, color, and orientation to mimic natural handwriting and scanning variability.  
Backgrounds include scanned paper textures and natural surfaces blended with random illumination, blur, and noise.

A dedicated synthetic generator was developed for SROIE-style forms to reproduce the visual and structural regularities of real receipts.  
Each template defines a spatial arrangement of key–value pairs such as vendor, date, total amount, and tax.  
Text values are instantiated from realistic random data (e.g., company names, addresses, numeric fields) and rendered with the same font pool and augmentation pipeline used for other synthetic pages.  
The generator outputs both the rendered RGB image and the exact bounding boxes of all textual fields, ensuring pixel-level alignment between content and annotations.  

\begin{table}[t]
\centering
\footnotesize
\caption{Datasets used for training and evaluation. HW = handwritten, P = printed.}
\label{tab:data_statistics}
\begin{tabularx}{\linewidth}{Xccccl}
\toprule
Dataset & Type & Train & Val & Test & Lang \\
\midrule
\multicolumn{6}{c}{\textit{Synthetic}}\\
IAM-synth & HW & 50\,000 & — & — & en\\
RIMES-synth & HW & 50\,000 & — & — & fr\\
SynthDOG & HW/P & 100\,000 & — & — & en/fr\\
SROIE-syn & P & 50\,000 & — & — & en\\[0.25em]
\multicolumn{6}{c}{\textit{Real}}\\
IAM & HW & 747 & 116 & 336 & en\\
RIMES & HW & 1\,050 & 100 & 100 & fr\\
SROIE’19 & P & 626 & — & 361 & en\\
MAURDOR & HW/P & 4\,515 & 742 & 742 & en/fr\\
IDL + PDFA & HW/P & 500\,000 & — & — & en\\
\bottomrule
\end{tabularx}
\end{table}
For MAURDOR, the counts reported in Table~\ref{tab:data_statistics} correspond to the French/English subset that we retained and re-annotated for our experiments, rather than to the full 8{,}129-document campaign release.
\subsection{Synthetic SROIE Generator}
We use a template–conditioned generator that renders receipt pages on a blank RGB canvas and provides line-level bounding boxes aligned with text. Base templates are derived from SROIE training annotations; before rendering, each template undergoes layout perturbations (field jitter, spacing and padding changes) to produce a novel instance. Content values are sampled from realistic distributions (company and contact information, addresses, dates, prices, identifiers), with the candidate closest in length to the target field selected to preserve line geometry. A font is drawn once per page from a curated pool of $\sim$4{,}000 families (serif/sans and handwriting), together with a size in the range 22–34\,px; we apply 50\% random uppercasing and light per-field rotations (1–6$^\circ$ with probability 0.2). Separator lines made from repeated ASCII symbols (\texttt{-}, \texttt{*}, \texttt{<}, \texttt{>}, \texttt{=}, \texttt{.}, \texttt{\_}) are stochastically inserted between blocks to mimic ornamental rules. Bounding boxes are recorded at paste time in pixels and quantised on a 10\,px grid to match PILOT’s coordinate tokenization. After compositing, we apply a light subset of degradations (illumination, blur/compression, ink bleed/mottling) to approximate scan/camera noise while maintaining legibility.

\subsection{Evaluation setup}
For handwritten English we use the IAM database with the standard RWTH writer-independent split, and for French handwriting we follow the official RIMES~2009 protocol. For printed receipts we use the SROIE~2019 dataset. In all cases we work at the page level: ground-truth line transcriptions are concatenated into a single string per page, and predictions are evaluated against this page-level reference. This avoids any dependence on a particular line-segmentation scheme and mirrors the full-page setting in which PILOT is intended to operate.

For text recognition, we report task-standard metrics for each dataset. On IAM and RIMES we use character error rate (CER) as the primary measure, and additionally compute BLEU and METEOR on the concatenated page text to complement CER with sequence-level similarity measures. BLEU evaluates the overlap of short $n$-grams between prediction and reference, and is therefore sensitive to local lexical accuracy and missing or spurious tokens. METEOR also compares prediction and reference at the token level, but balances precision and recall and is generally more tolerant to small word-order differences, which makes it useful for page-level OCR where minor reading-order variations may occur. On SROIE, where the official challenge focuses on word-level fields, we report a word-level F1 score in place of CER, and complement it with BLEU and METEOR over page-level text. Before computing any metric, we apply a shared normalization to both predictions and references: Unicode NFKC normalization, unification of curly quotes to straight quotes, collapse of all whitespace (including line breaks) to single spaces, and trimming of leading/trailing spaces. Layout markers and special tokens are removed prior to scoring. This design closely follows the spirit of the normalization used in recent HTR evaluations, which emphasize content over formatting, while still being simple enough to reproduce.\footnote{Exact normalization scripts will be released with the code.} Because evaluation is performed at the page level, our CER values are not directly comparable to line-level IAM/RIMES benchmarks, but they better reflect the full-document use cases that motivate our model.

Formally, let $u_i$ and $v_i$ denote the normalized ground-truth and predicted strings for page $i$. We compute CER as a micro-averaged Levenshtein distance,
\[
\mathrm{CER} = \frac{\sum_i \mathrm{edit}(u_i,v_i)}{\sum_i |u_i|},
\]
where $\mathrm{edit}(\cdot,\cdot)$ is the character-level edit distance. BLEU is computed at the page level using standard sentence-level BLEU with exponential smoothing, applied to the normalized strings; we average the resulting scores over all pages and report them as percentages. METEOR is computed per page using the NLTK implementation on tokenized normalized strings, and the page-level scores are averaged over the corpus. On SROIE, the reported F1 is a token-level word F1: for each page we treat the normalized ground-truth and predicted words as bags of tokens, measure the overlap in terms of token counts, and compute precision, recall, and F1; the final score is the mean F1 over all pages. BLEU and METEOR on SROIE are computed on the same normalized page strings as for IAM and RIMES; in practice BLEU is conservative on short, fragmented texts, so we primarily interpret SROIE performance through F1 and METEOR.

Closed-source models (\textit{GPT-4o}, \textit{GPT-4o-mini}, \textit{Claude-3-Haiku}) are accessed through their official APIs using greedy decoding (temperature~0) and otherwise default parameters. Open-source vision--language models (\textit{Qwen2.5-VL}, \textit{Qwen3-VL}, \textit{InternVL2.5}, \textit{Phi-3.5-Vision}, \textit{Nanonets-OCR}, \textit{OlmOCR}, \textit{PaddleOCR-VL}) are evaluated in zero-shot mode using the \texttt{vLLM} backend, with image resolution capped only by the model’s recommended maximum number of pixels.

To minimise prompt-induced variability, all VLMs share a common JSON-based transcription protocol. The system message constrains the model to act as a pure OCR engine on full-page document images: it must transcribe \emph{exactly} what is written (including accents, punctuation, casing and spacing), preserve line breaks and blank lines, follow a fixed reading order (columns left-to-right, lines top-to-bottom, tables row-by-row with single spaces between cells), and avoid any description, correction or normalization of the text. The user message specifies the task as ``transcribe the entire page at line level in the specified reading order’’ and requires the model to return \emph{only} valid JSON with the fixed schema
\texttt{\{"lines": ["line\_001\_text","line\_002\_text",...]\}}.
The exact system and user prompts used for all models are provided in Appendix~\ref{app:prompts}. When a model requires a specific chat template (e.g.\ special role tokens), we wrap these same texts inside the prescribed format while preserving their semantics and the JSON output schema. In all cases, decoding is deterministic (greedy or temperature~0) and limited to a single image per query to avoid batch-related artifacts.

Public VLMs rarely report results on IAM, RIMES~2009, or SROIE under our exact page-level setting and metric definitions. We therefore recompute all VLM baselines ourselves under this unified protocol. To reduce prompt-instability effects, we first run each model five times on a small calibration subset per dataset (20\%) while varying minor wording, and retain the template that yields the most stable CER. We provide the exact model identifiers (checkpoint tags/commits), prompts, and evaluation scripts, and mark all reproduced figures with \repro in the tables.

Classical end-to-end HTR systems such as DAN, Dessurt, and DANIEL are not reimplemented; instead we report their published numbers on IAM or RIMES whenever they use the same dataset variant (IAM with the RWTH split, RIMES~2009) and page- or paragraph-level recognition. 

\begin{table*}[t]
\centering
\renewcommand{\arraystretch}{1.15}
\setlength{\tabcolsep}{3.8pt}
\small
\caption{Cross-dataset OCR comparison (\%). Values marked with \repro\ were reproduced by us for VLMs under our evaluation protocol. BLEU is computed at the page level with sentence-level smoothing, and METEOR is averaged per page.}

\vspace{0.4em}

\begin{tabular}{l|ccc|ccc|ccc}
\toprule
\multirow{2}{*}{\textbf{Model}} &
\multicolumn{3}{c|}{\textbf{IAM (HW, EN)}} &
\multicolumn{3}{c|}{\textbf{RIMES (HW, FR)}} &
\multicolumn{3}{c}{\textbf{SROIE (P, EN)}}
\\
\cmidrule(lr){2-4}\cmidrule(lr){5-7}\cmidrule(lr){8-10}
& CER$\downarrow$ & BLEU$\uparrow$ & METEOR$\uparrow$
& CER$\downarrow$ & BLEU$\uparrow$ & METEOR$\uparrow$
& F1$\uparrow$ & BLEU$\uparrow$ & METEOR$\uparrow$
\\
\midrule
\multicolumn{10}{c}{\textbf{Traditional OCRs}}\\
\midrule
EasyOCR\repro~\cite{easyocr}	& 55.58	& 0.51	& 7.21 & 62.06 & 0.05 & 8.61	& 61.21 & 1.47 &	58.24\\
Tesseract\repro~\cite{smith2007overview}	& 50.93 & 3.04 & 10.09 & 73.95 & 1.26 & 8.14 &	67.77 & 7.19 & 67.99\\
\midrule
\multicolumn{10}{c}{\textbf{Closed-source VLMs}}\\
\midrule
GPT-4o-mini\repro~\cite{gpt4o} & 5.23 & 61.35 & 87.01 & 8.88 & 45.62 & 83.38 & 89.96 & 53.68 & 88.62 \\
Claude-3-Haiku\repro~\cite{claude} & 8.50 & 64.79 & 79.07 & 18.62 & 26.68 & 67.59 & 84.52 & 40.42 & 81.33 \\
GPT-4o\repro~\cite{gpt4o} & \best{2.20} & \best{85.55} & \best{91.36} & 5.13 & 66.15 & 86.39 & 92.04 & 58.32 & 91.65 \\
\midrule
\multicolumn{10}{c}{\textbf{Open-source VLMs}}\\
\midrule
Qwen2.5-VL-7B\repro~\cite{bai2025qwen2} & 3.66 & 76.96 & 86.83 & 6.66 & \secondbest{69.41} & 85.67 & 92.24 & 57.12 & 91.23 \\
Qwen3-VL-4B\repro~\cite{yang2025qwen3} & 3.27 & \secondbest{84.29} & 87.71 & 12.33 & 64.32 & 80.84 & 92.44 & \secondbest{58.39} & 91.56 \\
InternVL2.5-8B\repro~\cite{chen2024internvl} & 27.14 & 24.41 & 58.71 & 35.10 & 18.77 & 46.26 & 72.84 & 40.60 & 68.95 \\
Phi-3.5-Vision\repro~\cite{abdin2024phi} & 5.25 & 34.65 & 75.48 & 36.91 & 19.71 & 33.81 & 57.46 & 37.26 & 53.11 \\
Nanonets-OCR-s\repro~\cite{nanonetsocr} & 13.89 & 49.43 & 85.19 & 15.15 & 47.01 & 83.74 & 92.56 & 56.52 & \secondbest{92.23} \\
Nanonets-OCR2-3B\repro~\cite{nanonetsocr2} & \secondbest{2.83} & 83.01 & \secondbest{89.11} & 14.50 & 63.41 & \secondbest{88.68} & 76.30 & 43.65 & 79.17 \\
OlmOCR-2-7B\repro~\cite{poznanski2025olmocr} & 4.97 & 16.35 & 88.31 & 8.20 & 22.43 & 85.01 & 90.73 & 54.37 & 89.74 \\
PaddleOCR-VL\repro~\cite{cui2025paddleocr} & 6.28 & 34.29 & 84.42 & 14.85 & 27.09 & 67.49 & 88.23 & 51.05 & 87.46 \\
\midrule
\multicolumn{10}{c}{\textbf{End-to-End OCR Models}}\\
\midrule
DAN~\cite{coquenet2023dan} & 4.30 & - & - & \best{4.54} & - & - & - & - & - \\
Dessurt~\cite{davis2022end} & 4.80 & - & - & - & - & - & - & - & - \\
DANIEL~\cite{constum2025daniel} & 4.38 & - & - & 5.80 & - & - & - & - & - \\
ViTLP~\cite{mao2024visually} & - & - & - & - & - & - & \secondbest{92.79} & - & - \\
\midrule
PILOT & 4.31 & 77.06 & 86.99 & \secondbest{5.05} & \best{75.77} & \best{91.75} & \best{93.77} & \best{60.49} & \best{93.22} \\
\bottomrule
\end{tabular}
\label{tab:ocr_comparison}
\end{table*}

\noindent
Table~\ref{tab:ocr_comparison} highlights complementary trends across datasets. On English handwriting (\textsc{IAM}), \textbf{PILOT} remains competitive but trails larger or OCR-specialised models such as \textit{Nanonets-OCR2-3B} and \textit{GPT-4o}. This is consistent with the fact that our pretraining does not target the full diversity of IAM writers, whereas proprietary or specialised models may have seen similar distributions during training. In contrast, on French handwriting (\textsc{RIMES}), \textbf{PILOT} attains the best BLEU and METEOR, with a CER on par with the strongest VLMs, indicating that explicit spatial grounding can be beneficial in dense cursive layouts. On printed receipts (\textsc{SROIE}), \textbf{PILOT} achieves the highest F1 and sequence-level scores, outperforming both proprietary and open-source VLMs. BLEU is known to be conservative on short, fragmented texts such as receipts, so we primarily interpret SROIE performance through F1 and METEOR, for which PILOT is also strongest. Taken together, these results indicate that joint text--layout decoding provides consistent gains in structured or layout-dependent settings, while handwriting performance could further improve with targeted adaptation. Since our evaluation operates at page level with a unified normalization, absolute CER values on IAM and RIMES are not directly comparable to line-level HTR benchmarks, although the relative ordering of models is stable under reasonable variations of the metric.

\subsection{Text Detection Results}
We next assess line-level detection with \textsc{DetEval} (Precision/Recall/F1). For VLMs that do not natively return line boxes, we extract axis-aligned line-level bounding boxes from their outputs using the same post-processing as in the recognition pipeline. As reported in Table~\ref{tab:sroie_rimes_td}, \textbf{PILOT} attains the highest F1 on both \textsc{SROIE} and \textsc{RIMES}, outperforming classical detectors (CTPN, EAST) and recent VLM baselines where available. These gains mirror the recognition trends and indicate that the coordinate tokens trained in the unified decoder transfer well to stand-alone detection metrics, without requiring task-specific detection heads.

\begin{table}[t]
\centering
\caption{Text detection results on RIMES 2009 and SROIE.}
\begingroup
\fontsize{8}{10.5}\selectfont
\begin{tabularx}{0.85\linewidth}{l|ccc}
    \toprule
    \multicolumn{4}{c}{\textit{SROIE}}\\
    \midrule
    \textbf{Method} & \textbf{Precision} & \textbf{Recall} & \textbf{F1} \\
    \midrule
    Tesseract\repro~\cite{smith2007overview}           & 46.45 & 50.16 & 47.78 \\
    EasyOCR\repro~\cite{easyocr}             & 78.23 & 67.19 & 72.01 \\
    CTPN~\cite{ctpn}                & 81.14 & 87.23 & 84.07 \\
    EAST~\cite{zhou2017east}                & 85.07 & 87.17 & 86.11 \\
    ViTLP~\cite{mao2024visually}    & \secondbest{91.62} & \secondbest{91.68} & \secondbest{91.65} \\
    Qwen3-VL-4B\repro~\cite{yang2025qwen3}         & 70.69 & 41.69 & 44.38  \\
    \midrule
    PILOT     & \textbf{95.67} & \textbf{94.75} & \textbf{95.21} \\
    \midrule
    \multicolumn{4}{c}{\textit{RIMES}}\\
    \midrule
    Tesseract\repro~\cite{smith2007overview}           & 63.10 & 63.61 & 62.95 \\
    EasyOCR\repro~\cite{easyocr}             & 74.90 & 79.99 & 77.19 \\
    Qwen3-VL-4B\repro~\cite{yang2025qwen3}         & \secondbest{88.75} & \secondbest{92.32} & \secondbest{90.25} \\
    \midrule
    PILOT               & \textbf{96.86} & \textbf{94.83} & \textbf{95.67} \\
    \bottomrule
\end{tabularx}
\endgroup
\label{tab:sroie_rimes_td}
\end{table}

\subsubsection*{Heterogeneous benchmark: MAURDOR}
Beyond public benchmarks, we also evaluate PILOT on the MAURDOR benchmark~\cite{brunessaux2014maurdor}, a large-scale document analysis dataset collected for evaluating automatic processing of administrative and correspondence documents. We use the subset from the second evaluation campaign, which contains 8{,}129 heterogeneous documents written in French, English, and Arabic. MAURDOR is organized into five categories reflecting real-world document variability: C1 (printed forms to be completed by hand, 12\%), C2 (commercial documents such as invoices, quotations and leaflets, 40\%), C3 (private handwritten correspondence, 25\%), C4 (typed personal or professional correspondence, 20\%), and C5 (other content such as diagrams or drawings, 3\%). Even within a category, documents differ substantially in layout, background clutter, writing style, and the proportion of handwritten versus printed content, making MAURDOR a challenging full-page OCR benchmark.

We consider two complementary evaluation settings. First, following common practice in handwritten document recognition, we focus on categories C3 and C4, for which reading order can be reliably inferred from layout. Category C3 contains visually diverse handwritten letters and manuscripts, while C4 corresponds to more regular typed or mixed correspondence. As in the public benchmarks, evaluation is performed at page level. For C3 and C4, we report character error rate (CER), word error rate (WER), and line-level detection F1 computed with \textsc{DetEval}. Table~\ref{tab:maurdor_c3c4} summarizes these results.
Table~\ref{tab:maurdor_c3c4} summarizes these results.
\begin{table}[t]
\centering
\caption{\textbf{MAURDOR} (C3: private manuscripts, C4: professional correspondence). Recognition is evaluated with CER/WER; F1 is the \textsc{DetEval} text-detection F1 on the same pages.}
\begingroup
\fontsize{8}{10}\selectfont
\setlength{\tabcolsep}{1.5mm}
\begin{tabularx}{\linewidth}{X|ccc|ccc}
    \toprule
    & \multicolumn{3}{c|}{\textbf{C3}} & \multicolumn{3}{c}{\textbf{C4}} \\
    \textbf{Method} & CER $\downarrow$ & WER $\downarrow$ & F1 $\uparrow$ &
    CER $\downarrow$ & WER $\downarrow$ & F1 $\uparrow$ \\
    \midrule
    DAN\cite{coquenet2023dan}               & 8.62 & 18.94 & -     & \textbf{8.02} & 14.57 & -     \\
    PILOT              & \textbf{5.86} & \textbf{11.94} & \textbf{84.38} & 9.50 & \textbf{14.11} & 80.91 \\
    \bottomrule
\end{tabularx}
\endgroup
\label{tab:maurdor_c3c4}
\end{table}

\noindent
On the more challenging C3 subset, \textbf{PILOT} substantially outperforms DAN, reducing CER from $8.62$ to $5.86$ and WER from $18.94$ to $11.94$, while also achieving a detection F1 of $84.38$. This result is consistent with the stronger variability of C3 pages, which often contain less regular handwriting, noisier page scans, and more diverse layouts than standard correspondence benchmarks. On C4, which is visually more regular and closer to structured correspondence, the gap between systems narrows: DAN attains the lowest CER ($8.02$), whereas \textbf{PILOT} achieves a slightly lower WER ($14.11$ vs.\ $14.57$) together with a high detection F1 of $80.91$. Overall, these results suggest that PILOT is particularly advantageous when transcription and localization must remain robust under stronger document heterogeneity.

We next extend the analysis beyond C3/C4 and evaluate on a broader subset of MAURDOR covering all categories (C1–C5), restricted to pages containing French/English content. In this setting, we report page-level CER and METEOR and, when box predictions are available, line-level \textsc{DetEval} F1 on the same pages. Table~\ref{tab:maurdor_full} reports these results and confirms that the trends observed on C3/C4 remain consistent under a broader and more heterogeneous evaluation.

\begin{table}[t]
\centering
\caption{\textbf{MAURDOR} full handwritten and printed test set (all categories, French and English). CER and METEOR are recognition metrics; F1 is the \textsc{DetEval} text-detection F1 where available.}
\begingroup
\fontsize{8}{10}\selectfont
\setlength{\tabcolsep}{1.5mm}
\begin{tabularx}{0.9\linewidth}{l|ccc}
    \toprule
    \textbf{Method} & CER $\downarrow$ & METEOR $\uparrow$ & F1 $\uparrow$ \\
    \midrule
    GPT-4o-mini\repro~\cite{gpt4o}            & 6.89 & 73.16 & - \\
    GPT-4o\repro~\cite{gpt4o}                 & \secondbest{5.35} & \secondbest{87.12} & - \\
    Claude-3-Haiku\repro~\cite{claude}        & 17.11 & 59.92 & - \\
    Qwen2.5-VL-7B\repro~\cite{bai2025qwen2}   & 5.87 & 84.54 & \secondbest{83.16} \\
    Phi-3.5-Vision\repro~\cite{abdin2024phi}  & 15.66 & 61.03 & - \\
    PILOT                            & \best{5.23} & \best{89.67} & \best{90.10} \\
    \bottomrule
\end{tabularx}
\endgroup
\label{tab:maurdor_full}
\end{table}

\noindent
Across the French/English MAURDOR subset, \textbf{PILOT} attains the lowest CER ($5.23$) and the highest METEOR ($89.67$), slightly outperforming GPT-4o and Qwen2.5-VL-7B in recognition quality. Among models for which detection outputs are available, \textbf{PILOT} also achieves the best detection F1 ($90.10$ vs.\ $83.16$ for Qwen2.5-VL-7B). These results indicate that the coordinate-aware decoder generalizes beyond curated public benchmarks and remains effective on heterogeneous real administrative and correspondence documents.

\subsection{Prompt-Controlled Generation Tasks}
\subsubsection*{Fine-grained OCR}

The ability to restrict text recognition to user-specified areas of a document has become increasingly relevant with the emergence of large multimodal models capable of region-conditioned reasoning. Recent architectures such as FoX, mPLUG-DocOwl\,1.5, and GOT-OCR\,2.0 integrate spatial cues directly into their generative interface, enabling fine-grained extraction of textual content without the need for a separate detection module. However, their large parameter counts make them computationally heavy and difficult to deploy in real-world pipelines. This motivates the design of lighter architectures that retain fine-grained controllability while preserving recognition accuracy.

In this context, we evaluate the proposed \textbf{PILOT} model on region-level and line-level OCR tasks, where the model must transcribe only the content within a designated bounding box. These experiments assess whether the coordinate tokens introduced in our unified decoder allow accurate spatial grounding between the input region and the generated text.

As shown in Table~\ref{tab:fine_grained_ocr}, \textbf{PILOT} achieves the lowest edit distance (0.038) and the highest F\textsubscript{1} score (0.978) among all compared systems, surpassing both FoX and GOT-OCR\,2.0 on region-level recognition. The model maintains a balanced precision–recall trade-off, demonstrating that its coordinate-aware decoder effectively constrains generation to the targeted spatial region. BLEU and METEOR remain comparable to those of GOT-OCR\,2.0, indicating that lexical fluency and local coherence are preserved despite the smaller language head.

\begin{table}[t]
    \centering
    \caption{Comparison on fine-grained OCR on English documents.}
    \subfloat[{\textbf{Region level}}]{
    \begingroup
    \fontsize{8}{10.5}\selectfont
    \begin{tabularx}{\linewidth}{l|cccc}
        \toprule
        \textbf{Method} &        DocOwl1.5\cite{hu2024mplug} & Fox\cite{liu2024focus}  & GOT\cite{wei2024general}    & PILOT\\
        \midrule
        Edit $\downarrow$ & 0.435 & 0.059 & 0.041 & \best{0.038}\\
        F1 Score $\uparrow$        & 0.670 & 0.957 & 0.972 & \best{0.978}\\
        Precision $\uparrow$       & 0.886 & 0.962 & 0.973 & \best{0.982}\\
        Recall $\uparrow$          & 0.617 & 0.955 & \best{0.969} & \best{0.969}\\
        BLEU $\uparrow$            & 0.478 & 0.914 & \best{0.926} & 0.913\\
        METEOR $\uparrow$          & 0.569 & 0.955 & 0.966 & \best{0.972}\\
        \bottomrule
    \end{tabularx}
    \endgroup
    }

    \subfloat[{\textbf{Line level}}]{
        \begingroup
        \fontsize{8}{10.5}\selectfont
        \begin{tabularx}{0.62\linewidth}{l|cccc}
            \toprule
            \textbf{Method} &          Fox~\cite{liu2024focus} & PILOT\\
            \midrule
            Edit Distance $\downarrow$ &  0.116 & \best{0.053}  \\
            F1 Score $\uparrow$        &  0.879 & \best{0.921} \\
            Precision $\uparrow$       &  0.879 & \best{0.927}  \\
            Recall $\uparrow$          &  0.883 & \best{0.918}  \\
            BLEU $\uparrow$            &  0.845 & \best{0.879} \\
            METEOR $\uparrow$          &  0.878 & \best{0.905}  \\
            \bottomrule
        \end{tabularx}
        \endgroup
    }

    \label{tab:fine_grained_ocr}
\end{table}

On line-level evaluation, \textbf{PILOT} continues to deliver strong performance, reducing the edit distance from 0.116 to 0.053 and improving F\textsubscript{1} from 0.879 to 0.921 relative to FoX. These gains indicate that the proposed architecture maintains high character-level precision and structural consistency even when applied to fine-grained text segments. BLEU and METEOR scores remain close across both models, suggesting that improvements primarily stem from low-level transcription accuracy rather than higher-order linguistic modelling.

Overall, these experiments show that precise spatial grounding can be achieved without relying on very large multimodal backbones. The unified coordinate–text decoder of \textbf{PILOT} provides accurate region control and competitive transcription quality while remaining substantially lighter than existing document-oriented VLMs. This supports the view that spatial reasoning and OCR can be effectively integrated in a single, compact generative framework.

\begin{figure}[t]
\centering
\includegraphics[width=0.7\linewidth]{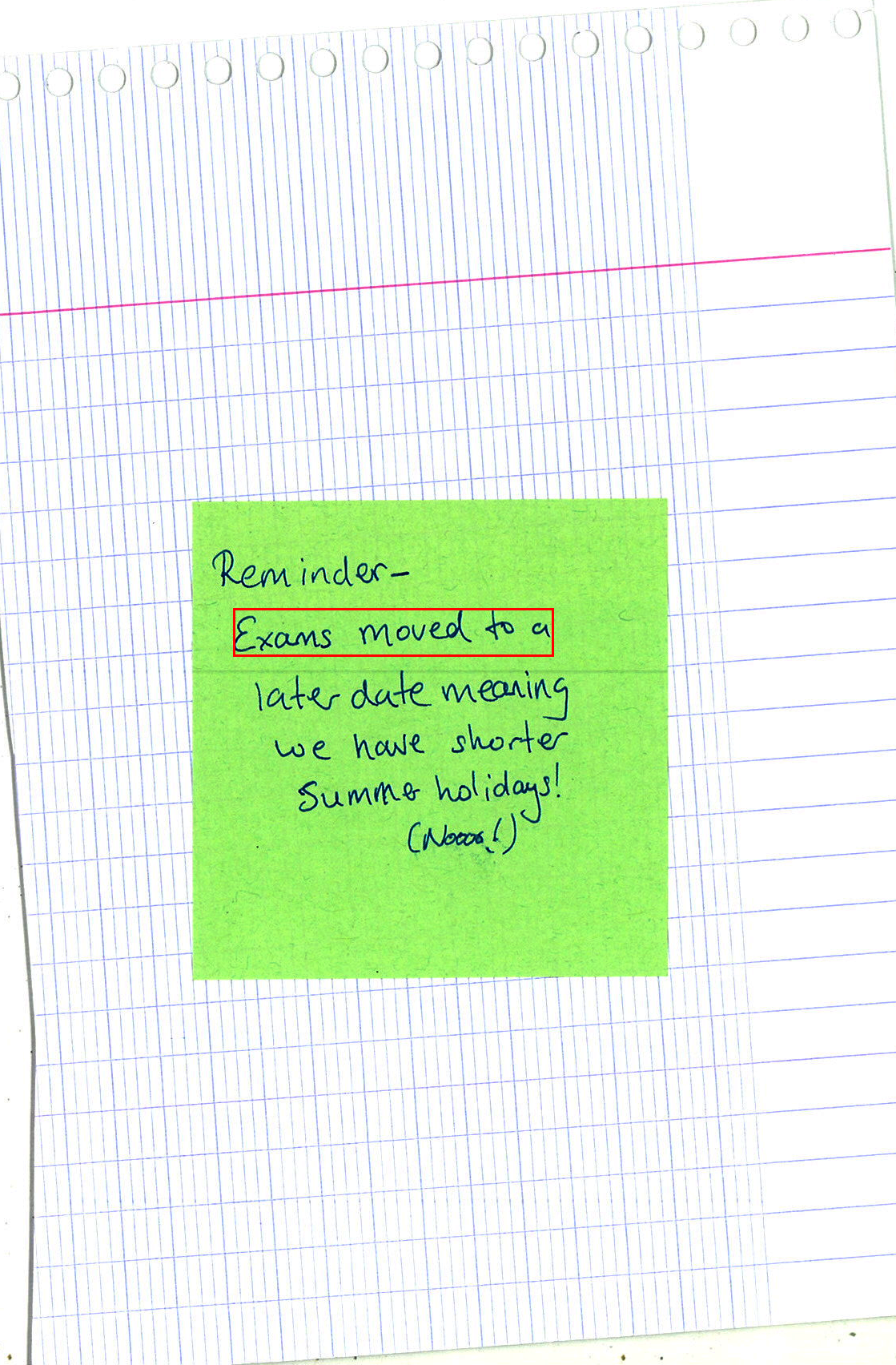}
\caption{Fine-grained OCR on MAURDOR sample.}
\vspace{2pt}%
{\scriptsize
        \textbf{Prompt:} ocr\_on\_box $31,139,71,145$\\
        \textbf{Label:}'exams moved to a'\\
        \textbf{Prediction:}'evans moved to a'
    
}
\label{fig:ocr_on_box_samples}
\end{figure}

Qualitative examples of region-conditioned OCR are shown in Figure~\ref{fig:ocr_on_box_samples}. The prompt token \texttt{<ocr\_on\_box>} is followed by the absolute pixel coordinates of a rectangular region. PILOT is asked to transcribe only the text inside this box, without hallucinating surrounding content.

\subsubsection*{Query-by-String Spotting}
\paragraph{\textbf{Against HTR spotters}}
We first evaluate \textbf{PILOT} on query-by-string (QbS) word spotting, a task where the model must localize textual instances that match a given query without explicit region annotations. Since none of the pretraining datasets include word-level bounding boxes, PILOT is initially trained for line-level spotting and later fine-tuned on the IAM handwriting dataset to adapt to the word-level QbS setting. The evaluation follows the standard protocol, where stop words are removed from the query set, mis-segmented instances are discarded, and tokens are lowercased for word-level experiments, while line-level evaluation remains case-sensitive.

As shown in Table~\ref{QbS}, \textbf{PILOT} achieves state-of-the-art performance with mean average precision (mAP) of 89.8\,\% and 88.9\,\% at IoU thresholds of 25\,\% and 50\,\%, respectively. These results surpass prior dedicated HTR spotters such as Ctrl-F-Net~(PHOC/DCToW) and SegFreeKWS, confirming that the unified coordinate–text decoder produces bounding boxes that are both spatially precise and semantically coherent with their corresponding transcripts. These gains indicate that the unified text--coordinate decoding scheme is sufficient to learn accurate associations between textual content and bounding-box predictions, without requiring a separate spotting-specific embedding objective or retrieval loss.

\begin{table}[t]
    \centering
    \caption{Results for the QbS word-spotting experiments (mAP\,\%) on IAM dataset.}
    \begingroup
    \fontsize{8}{10.5}\selectfont
    \begin{tabularx}{0.65\linewidth}{l|cc}
        \toprule
        \textbf{Method} & 25\% & 50\% \\
        \midrule
         BG index~\cite{ghosh2015query} &           -   &  48.6 \\
         Ctrl-F-Net DCToW~\cite{ctrlfnet} &  82.5 & 80.3  \\
         Ctrl-F-Net PHOC~\cite{ctrlfnet}   &  80.8 & 78.8  \\
         SegFreeKWS~\cite{retinas2023} & 85.8 & 59.2 \\
         \hline
         PILOT & \best{89.8} & \best{88.9} \\
        \bottomrule
    \end{tabularx}
    \endgroup
    \label{QbS}
\end{table}

\paragraph{\textbf{Against a pipelined system}}
To further validate the efficiency of PILOT as an end-to-end system, we compare it against a traditional two-stage pipeline composed of perfect word localization followed by text recognition using TrOCR-Base trained on cropped handwritten regions. This setting represents an oracle baseline for localization, since the word bounding boxes are derived directly from ground truth annotations.

At inference time, all ground-truth word boxes from the IAM test split are cropped and transcribed independently with TrOCR. Query-by-string retrieval is then performed by exact string matching over the recognized transcripts: for each query, we select all predicted word instances whose recognized text matches the query (strict setting) or satisfies CER\,$\leq$\,0.2 (relaxed setting). No embedding-based retrieval or similarity scoring is used; ranking metrics are computed directly from these matched instances using their bounding boxes and confidence scores.

Evaluation is conducted on the IAM handwriting test split, which contains 11\,306 unique non-stop-word queries (minimum length three characters). A prediction is counted as correct when its generated text matches the query and the predicted bounding box overlaps the ground-truth region with an IoU\,$\geq$\,0.5.

As reported in Table~\ref{tab:qbs_pipeline}, \textbf{PILOT} consistently outperforms the TrOCR-based pipeline under both matching criteria. Although the pipeline benefits from oracle localization, its performance remains strongly dependent on transcription errors and cannot recover missed instances. In contrast, PILOT jointly models localization and recognition at page level, which leads to substantially higher recall and ranking metrics even under exact matching.

\begin{table}[t]
  \centering
  \caption{Word-spotting results on IAM against a pipelined oracle baseline. The \emph{strict} setting requires exact text matching, while the \emph{relaxed} setting allows CER\,$\leq$\,0.2.}
  \label{tab:qbs_pipeline}
  \begingroup
  \fontsize{8}{10.5}\selectfont
  \setlength{\tabcolsep}{3.5pt}
  \begin{tabular}{llcccc}
    \toprule
    \textbf{Setting} & \textbf{System} & \textbf{mAP}$\uparrow$ & \textbf{P@10}$\uparrow$ & \textbf{R@1}$\uparrow$ & \textbf{Recall}$\uparrow$ \\
    \midrule
    \multirow{2}{*}{Strict}
      & PILOT & \best{0.8772} & \best{0.8895} & \best{0.8996} & \best{0.8878} \\
      & TrOCR~\cite{li2023trocr} & 0.4658 & 0.5635 & 0.5661 & 0.4708 \\
    \midrule
    \multirow{2}{*}{Relaxed}
      & PILOT & \best{0.8870} & \best{0.8890} & \best{0.8990} & \best{0.8870} \\
      & TrOCR~\cite{li2023trocr} & 0.6130 & 0.7280 & 0.7310 & 0.6590 \\
    \bottomrule
  \end{tabular}
  \endgroup
\end{table}
\paragraph{\textbf{Latency in the QbS setting}}
Finally, latency measurements on an RTX\,5000 Ada GPU (mixed precision, beam size\,=\,5) confirm the computational benefits of our approach in the query-by-string setting, see Table~\ref{tab:qbs_latency}. For all 11\,306 queries in the IAM test split, PILOT completes inference in 5\,521\,s (2.05 queries/s), compared with 6\,919\,s (1.63 queries/s) for TrOCR-Base. Despite performing detection and recognition jointly, PILOT achieves higher throughput while requiring less than half the parameters (155\,M vs.\ 334\,M), showing that, in this setting, an integrated generative decoder can be both faster and more accurate than a larger modular OCR pipeline.

\begin{table}[t]
\centering
\caption{Latency comparison in the query-by-string setting on the IAM test split (11\,306 queries). Measurements are reported on an RTX\,5000 Ada GPU with mixed precision and beam size\,=\,5.}
\label{tab:qbs_latency}
\begingroup
\fontsize{9}{10.5}\selectfont
\setlength{\tabcolsep}{3.5pt}
\begin{tabular}{lccc}
    \toprule
    \textbf{System} & \textbf{Time}$\downarrow$ & \textbf{Throughput}$\uparrow$ & \textbf{Params} \\
    \midrule
    PILOT & \best{5\,521\,s} & \best{2.05\,q/s} & \best{155\,M} \\
    TrOCR~\cite{li2023trocr} & 6\,919\,s & 1.63\,q/s & 334\,M \\
    \bottomrule
\end{tabular}
\endgroup
\end{table}

\subsubsection*{Full-page OCR runtime against VLM baselines}
To complement the query-by-string latency reported above, we also measure full-page OCR throughput of representative VLM baselines and \textbf{PILOT} on the combined IAM+RIMES test sets (436 pages). All experiments are run on a single NVIDIA A100\,80\,GB GPU with batch size~1, greedy decoding (temperature~0), and a cap of 1{,}600 generated tokens per page, using the same JSON-based transcription prompt as in Section~\ref{sec:experiments}. VLMs are served through the \texttt{vLLM} runtime with paged attention and fused kernels,
while PILOT is implemented as a plain PyTorch model without additional serving optimizations. As summarised in Table~\ref{tab:runtime}, the VLMs process between 0.63 and 0.89 pages/s (about 1.1--1.6\,s per page), whereas \textbf{PILOT} reaches 2.30 pages/s (about 0.43\,s per page) under the same conditions. Despite having one order of magnitude fewer parameters than the 3--7\,B VLMs, PILOT therefore offers a clearly favourable accuracy--latency trade-off for full-page OCR, and we expect further speedups to be possible with a dedicated serving stack or light quantization.

\begin{table}[t]
\centering
\caption{Runtime comparison on the IAM+RIMES test set (436 pages) on a single NVIDIA A100~80\,GB, batch size~1. Throughput is computed as number of pages divided by wall-clock time, including image loading and JSON decoding, with greedy decoding (temperature~0) and a maximum of 1{,}600 output tokens per page.}
\label{tab:runtime}
\begingroup
\fontsize{8}{10}\selectfont
\setlength{\tabcolsep}{1.7mm}
\begin{tabular}{l|c}
\toprule
\textbf{Model} & \textbf{Throughput (pages/s)} $\uparrow$ \\
\midrule
Qwen3-VL-4B\repro & 0.79 \\
Qwen2.5-VL-7B\repro & 0.64 \\
OlmOCR-2-7B\repro & 0.63 \\
Nanonets-OCR-s\repro & 0.89 \\
Nanonets-OCR2-3B\repro & 0.75 \\
Phi-3.5-Vision\repro & 0.87 \\
\midrule
PILOT & \textbf{2.3} \\
\bottomrule
\end{tabular}
\endgroup
\end{table}

\section{Ablation study}
\subsection*{Relative vs Absolute coordinates}
\label{relative_vs_absolute}
To assess how coordinate encoding affects performance, we compare the full model trained with our default \emph{absolute} 10-pixel grid (Absolute) against an otherwise identical version that predicts \emph{relative} coordinates (Relative). Both variants are trained with the same schedule, joint text--layout objective, data, and hyperparameters.
\begin{table}[t]
    \centering
    \caption{Text detection (TD) and text recognition (TR) results on
    \textsc{SROIE}.}    
    \begingroup
    \fontsize{9}{10.5}\selectfont
    \begin{tabularx}{0.75\linewidth}{l|ccc}
        \toprule
        \textbf{Method} & F1$_{\textit{TR}}$ & F1$_{\textit{TD}}$ & mAP$_{0.50{:}0.95}$\\
        \midrule
        Absolute             & 93.90 & 95.21 & 94.15\\
        Relative             & 93.67 & 87.32 & 85.97\\
        \bottomrule
    \end{tabularx}
    \endgroup
    \label{absolute_relative}
\end{table}

Table~\ref{absolute_relative} shows that absolute encoding yields consistent gains, most notably for text detection (+7.8 F1). We hypothesize that fixed absolute tokens align more naturally with the convolutional feature grid and are therefore easier for the decoder to learn, whereas relative offsets introduce additional complexity because they depend on the input image resolution. Most of the prediction errors we observe correspond to misalignments between text and boxes.
\begin{figure}[t]
    \centering
    \includegraphics[width=\linewidth]{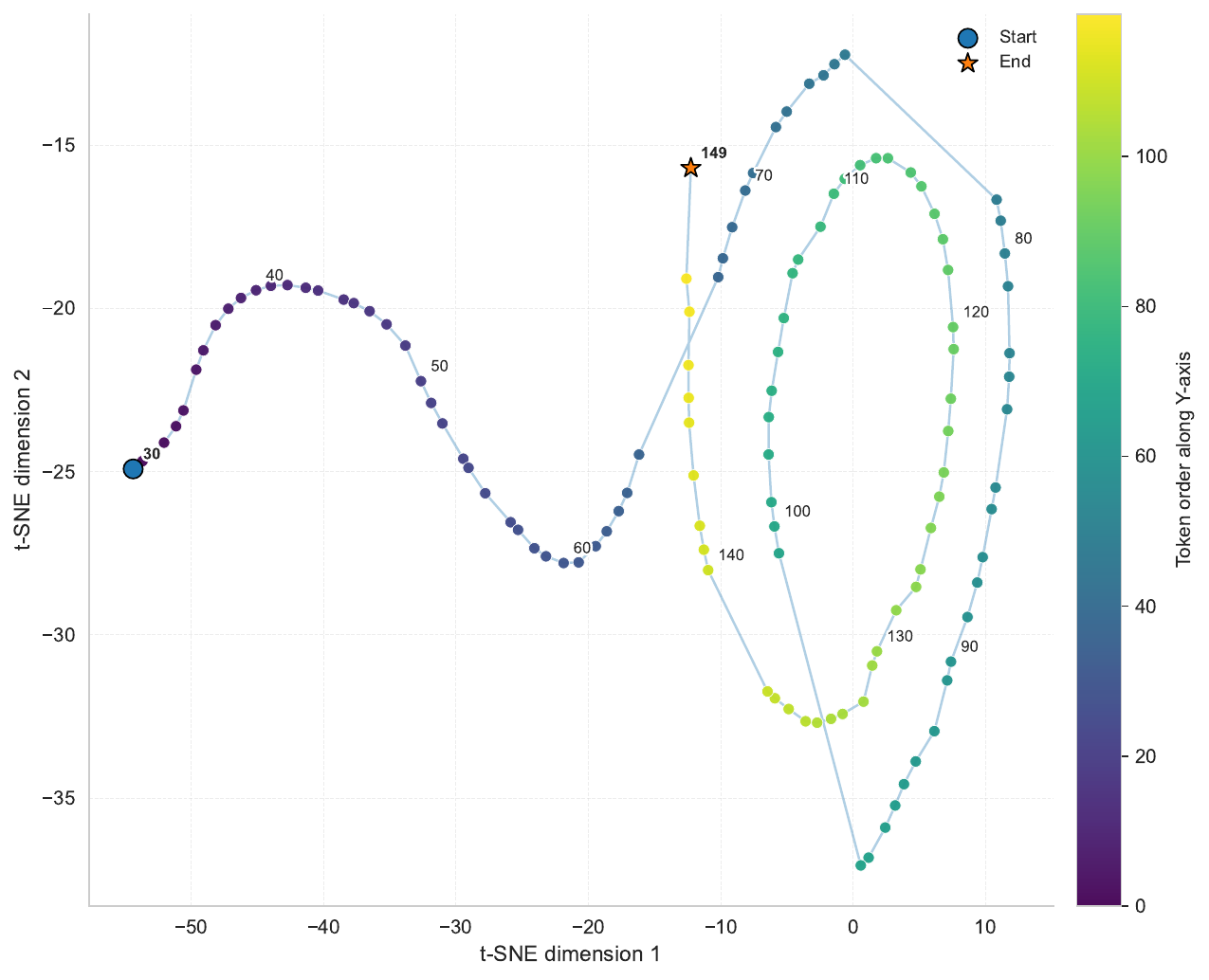}
    \caption{t-SNE projection of the learned \textit{Y}-axis coordinate-token embeddings. Tokens representing neighboring rows in the document cluster along a smooth curve, reflecting the model’s internalized notion of vertical order.}
    \label{positiona_tok_embed}
\end{figure}
The embedding geometry supports this interpretation. As visualized in Fig.~\ref{positiona_tok_embed}, tokens corresponding to nearby absolute positions form a coherent manifold, indicating that the model has learned a precise and monotonic mapping from tokens to embedding space.

\subsection*{One vs.\ Two Branches Approach}
To test whether explicit coordinate regression is necessary, we compare our default \emph{single-branch} model which emits absolute-grid \texttt{10\,px} tokens against a \emph{dual-branch} variant that attaches a lightweight MLP head ($1024\!\!\to\!\!512\!\!\to\!\!4$) to regress continuous $(x_1,y_1,x_2,y_2)$ offsets whenever a special \texttt{<loc>} token is generated. Both variants share the same 155 M-parameter backbone; the regression head adds only~1.5 M parameters. Coordinates are first normalized to $[0,1]$ and the head is optimised with the G-IoU loss~\cite{rezatofighi2019generalized}. Training is identical to the original version except for an additional warm-up phase in which only the MLP head is updated. We evaluate on the complete MAURDOR test partition. Results are summarised in Table \ref{tab:one_vs_two_branches}.
\begin{table}[t]
\centering
\caption{Single- vs.\ dual-branch localization on MAURDOR (all categories).  All metrics are percentages.  mAP follows COCO $[\text{IoU}=0.50{:}0.95]$.}
\label{tab:one_vs_two_branches}
\begingroup
\fontsize{9}{10.5}\selectfont
\begin{tabularx}{0.8\linewidth}{lcc}
\toprule
\textbf{Metric} & \textbf{PILOT} & \textbf{PILOT  + MLP} \\
\midrule
mAP$_{0.50{:}0.95}$ & 90.11 & 89.70 \\
AP$_{50}$           & 95.7 & 96.7 \\
AP$_{75}$           & 91.0 & 90.2 \\
AP$_{90}$           & 68.0 & 65.4 \\
F1 score            & 92.1 & 91.8 \\
\bottomrule
\end{tabularx}
\endgroup
\end{table}
The single-branch grid model slightly outperforms the regression variant on overall mAP (+0.4 pp) and high-IoU accuracy (AP$_{90}$), while being simpler and 4 ms/page faster at inference. This outcome is \emph{consistent with the annotation noise} in MAURDOR: perturbing each ground-truth box by only $\pm5$ px already lowers the mean IoU to 0.92, whereas snapping the same boxes to our \texttt{10\,px} grid lowers it to 0.90. In other words, the grid quantization error is below the human noise floor; refining boxes further yields little benefit but adds optimisation complexity. Similar conclusions were reported for Pix2Seq~\cite{chen2022pix2seq,chen2022pix2seqv2} showing that discrete coordinate tokens can match or surpass continuous regression when label noise exceeds a few pixels.

\subsection*{Encoding strategies for text and spatial tokens}
A generative OCR model must decide \emph{(i)} whether $x$- and $y$ coordinates share the same vocabulary and \emph{(ii)} how to interleave location tokens with the transcription.
We therefore evaluate four plausible text--location encoding schemes:
\begin{enumerate}
    \item Scenario: $[x_1]\,[y_1]\; w_1\,w_2\,\dots\,w_k\; [x_2]\,[y_2]$
    \item Scenario: $[loc_1]\,[loc_1]\; w_1\,w_2\,\dots\,w_k\; [loc_2]\,[loc_2]$
    \item Scenario: $[x_1]\,[y_1]\,[x_2]\,[y_2]\; w_1\,w_2\,\dots\,w_k$
    \item Scenario: $w_1\,w_2\,\dots\,w_k [x_1]\,[y_1]\,[x_2]\,[y_2]\;$
\end{enumerate}
Here, $x_i$, $y_j$, and $loc_k$ denote location tokens, while $w_k$ denotes a text token. 
\vspace{-0.5em}
\begin{table}[t]
\centering
\caption{Detection performance of PILOT on MAURDOR (all categories) under different encoding schemes. Metrics are percentages.}
\label{tab:encodings_comparison}
\begingroup
\fontsize{9}{10.5}\selectfont
\begin{tabularx}{0.8\linewidth}{lccc}
\toprule
\textbf{Encoding} & \textbf{Precision} & \textbf{Recall} & \textbf{F1} \\
\midrule
Scenario $_\text{1}$ & 95.9 & 94.3 & 95.1 \\
Scenario $_\text{2}$ & 95.1 & 94.0 & 94.6 \\
Scenario $_\text{3}$ & 95.8 & 94.2 & 95.0 \\
Scenario $_\text{4}$ & 95.7 & 94.1 & 94.9 \\
\bottomrule
\end{tabularx}
\endgroup
\end{table}

Using \emph{separate} vocabularies for the $x$- and $y$-axes (S1, S3, S4) provides a consistent $\sim$0.5\,pp F1 gain over a single, shared location vocabulary (S2), while adding a negligible number of classes compared to the 34 k text tokens, so inference speed is unchanged. The \emph{ordering} of text and coordinates, on the other hand, has no measurable impact; the model learns to associate words with their bounding boxes equally well whether coordinates come before, after, or split around the transcription.  We therefore adopt the S1 format, separate $x$/$y$ tokens with top-left coordinates emitted first, as our default because it is slightly more accurate and keeps a natural reading order.

\subsection*{Impact of grid quantization}
The default PILOT configuration discretises each page into \texttt{10\,px} cells. To estimate the benefit of finer localization we re-train two additional variants that use \texttt{5\,px} and \texttt{2\,px} location units; nothing else in the architecture changes. Reducing the cell size enlarges the \emph{location} vocabulary (Table\ref{tab:quantization_precision_maurdor}) and lengthens the generated sequence, but introduces no extra parameters.

\begin{table}[t]
    \centering
    \caption{Text-detection accuracy on the MAURDOR test set for different grid resolutions.  mAP is averaged over IoU thresholds 0.50–0.95.}
    \label{tab:quantization_precision_maurdor}
    \begingroup
    \fontsize{9}{10.5}\selectfont
    \begin{tabularx}{0.8\linewidth}{c|cc}
        \toprule
        \textbf{Pixels / Unit} & \textbf{F1 score} & \textbf{mAP$_{0.50{:}0.95}$} \\
        \midrule
        10 px & 92.1 & 90.11 \\
         5 px & 92.8 & 91.0 \\
         2 px & 92.9 & 91.3  \\
        \bottomrule
    \end{tabularx}
    \endgroup
\end{table}

Halving the grid step from 10 px to 5 px raises mAP by 0.9 pp and F1\textsubscript{full} by 0.7 pp, showing that most residual localization error stems from coarse quantization. Further refining to 2 px adds only another 0.3 pp mAP while more than doubling the sequence length, resulting in longer decoding times ($\approx 0.5$ s per 100 tokens on an NVIDIA RTX 5000 16~GB GPU). This saturation mirrors findings by Pix2Seq for COCO detection: once the cell size drops below the annotation-noise floor ($\approx \pm5$ px on MAURDOR), discrete bins perform on par with continuous regression while keeping the single-branch design intact. Consequently, we adopt the 10 px grid for the speed–accuracy trade-off and treat the 5 px variant as an optional high-precision setting.

\subsection*{Effect of the three-stage curriculum}
\label{subsec:curriculum_ablation}

The training schedule in Section~\ref{subsec:training_curriculum} introduces text, layout, and prompt-conditioned objectives in three successive stages. To quantify its impact, we compare four training strategies, all using the same data, optimizer and hyper-parameters, but differing in the order in which tasks are introduced.

\noindent\textbf{Strategy A}: Single-stage multi-task training from scratch on all objectives at once: page-level OCR, joint text+box generation, and prompt-controlled tasks.

\noindent\textbf{Strategy B}: Two-stage schedule without plain transcription pretraining. Training starts directly with joint text+box prediction and then adds prompt-controlled tasks.

\noindent\textbf{Strategy C}: Two-stage schedule without prompt-controlled training. The model is trained first on plain OCR, then on joint text+box prediction, and is never exposed to prompt-conditioned examples.

\noindent\textbf{Strategy D}: Full three-stage curriculum as described in Section~\ref{subsec:training_curriculum}, with plain OCR pretraining, joint text+box generation, and a final prompt-controlled stage.

We evaluate these strategies on the SROIE validation set. For text recognition we report page-level word F1 and METEOR, using the same normalization as in Section~\ref{sec:experiments}. For spatial reasoning we report line-level detection F1 computed with \textsc{DetEval}. Table~\ref{tab:curriculum_ablation} summarises the results.\footnote{All models share the same architecture and training budget; only the task schedule is changed.}

\begin{table}[t]
    \centering
    \caption{Impact of the training curriculum on SROIE (validation set). Recognition F1 and METEOR are computed at page level; detection F1 is line-level \textsc{DetEval} F1. All values are percentages.}
    \label{tab:curriculum_ablation}
    \begingroup
    \fontsize{8}{10}\selectfont
    \setlength{\tabcolsep}{1.8mm}
    \begin{tabular}{l|ccc}
        \toprule
        \textbf{Training strategy} &
        F1$_\text{rec}$ $\uparrow$ & METEOR$_\text{rec}$ $\uparrow$ & F1$_\text{det}$ $\uparrow$ \\
        \midrule
        Strategy A & 92.9 & 92.5 & 90.8 \\
        Strategy B & 93.5 & 92.9 & 92.2 \\
        Strategy C & 93.4 & 92.9 & 94.1 \\
        \textbf{Strategy D} & \best{93.6} & \best{93.2} & \best{95.6} \\
        \bottomrule
    \end{tabular}
    \endgroup
\end{table}

All strategies reach broadly similar text recognition quality on SROIE: the gap between the worst and best configuration is below one point in page-level F1 and METEOR. In other words, a model trained directly on all objectives (Strategy A) or without plain OCR pretraining (Strategy B) still learns to transcribe receipts reasonably well.

The differences are much larger once spatial reasoning is involved. The single-stage multi-task baseline in Strategy A underperforms Strategy D by almost 5 points in detection F1, and often produces boxes that are systematically misaligned with the underlying text. Strategy B recovers part of this gap by dedicating a phase to joint text+box learning, but its detection accuracy remains clearly below that of Strategy D, indicating that the decoder struggles to align coordinates and content when it has not first internalised a stable language model over page text.

Strategy C, which omits the prompt-controlled stage but follows the plain-OCR then text+box schedule, performs almost as well as Strategy D on page-level recognition and is only moderately worse on detection F1. However, when evaluated on region-based OCR and query-by-string tasks (Table~\ref{tab:fine_grained_ocr} and Table~\ref{QbS}), this strategy degrades noticeably: the model tends to drift outside the requested region or to truncate its answers when conditioned on explicit spatial prompts.

By contrast, Strategy D combines the benefits of all phases. Plain OCR pretraining stabilises learning and speeds up convergence; the joint text+box phase teaches consistent alignment between words and coordinates; and the final prompt-controlled stage is crucial for robust behaviour under region and content prompts. Together with the architectural ablations that follow, these results support the conclusion that both the unified decoder and the progressive three-stage curriculum are necessary to obtain strong spatial grounding while preserving high text recognition performance.

\subsection*{Impact of encoder attention modules}
The visual encoder of PILOT is built from convolutional and depthwise-separable convolutional blocks, augmented with Efficient Channel Attention (ECA) and CBAM spatial attention. To assess whether these modules are actually useful, we compare three encoder variants under the same training schedule and data: (i) a plain convolutional backbone without attention, (ii) the same backbone with ECA only, and (iii) the full encoder with both ECA and CBAM. We report page-level recognition and line-level detection on the SROIE validation set, which provides a representative structured OCR benchmark with reliable line-box annotations.

\begin{table}[t]
\centering
\caption{Ablation of encoder attention modules on SROIE validation. Recognition is evaluated at page level; detection is line-level \textsc{DetEval} F1.}
\label{tab:encoder_ablation}
\begingroup
\fontsize{8}{10}\selectfont
\setlength{\tabcolsep}{2.5mm}
\begin{tabular}{l|ccc}
\toprule
\textbf{Encoder variant} & F1$_\text{rec}$  & METEOR$_\text{rec}$  & F1$_\text{det}$  \\
\midrule
CNN               & 93.76 & 93.28 & 94.05 \\
CNN + ECA         & 93.70 & 93.13 & 94.89 \\
CNN + ECA + CBAM  & 93.77 & 93.22 & 95.21 \\
\bottomrule
\end{tabular}
\endgroup
\end{table}

Table~\ref{tab:encoder_ablation} shows that adding encoder attention has only a marginal effect on page-level recognition, with F1$_\text{rec}$ and METEOR$_\text{rec}$ remaining nearly unchanged across all variants. By contrast, detection benefits more clearly from these modules: ECA alone improves line-level F1 from 94.05 to 94.89, and adding CBAM further raises it to 95.21. This suggests that channel and spatial attention primarily improve the encoder's ability to emphasize text-bearing regions and suppress background clutter, which is more beneficial for localization than for transcription itself. We therefore retain the full encoder in PILOT because it improves spatial grounding while preserving recognition quality.
\section{Discussion}
Although \textbf{PILOT} delivers strong end-to-end performance in both text recognition and detection, the current work is deliberately focused on \emph{pure} OCR scenarios: full-page transcription, line-level detection, region-based reading, and query-by-string spotting. The model always decodes sequences of text tokens interleaved with line-level boxes and is evaluated primarily with sequence-level metrics (CER, METEOR, F1) and detection scores. In its present form, PILOT does not attempt to perform higher-level document understanding tasks such as named entity recognition, key–value extraction, table structure recovery, or visual question answering. These tasks are typically addressed by larger multimodal LLMs that map images directly to structured JSON or free-form answers. We view PILOT as complementary to this line of work: it provides a compact, spatially grounded OCR core that could be coupled with a downstream language head, but we leave such integration and joint training for future work.

\section{Conclusion}
We introduced \textbf{PILOT}, a lightweight generative framework that unifies text detection, recognition, and spatial understanding within a single Transformer decoder. By jointly generating text and spatial tokens in a unified decoding stream, our approach eliminates error-prone cascaded processing while enabling unprecedented prompt-based control for tasks like region-specific reading and content localization. PILOT achieves competitive accuracy across diverse handwritten and printed document benchmarks despite its compact architecture. The model’s efficiency and flexibility, validated through extensive experiments, demonstrate that sophisticated spatial reasoning does not require massive models. Future work will extend this prompt-driven paradigm to table parsing and visual question answering.

\section*{Acknowledgements}
This work was granted access to the HPC resources of CRIANN (Regional HPC Center, Normandy, France) and GENCI-IDRIS. The authors gratefully acknowledge this computational support.

\appendix
\section{Prompts used for VLM evaluation}
\label{app:prompts}
\paragraph{Text recognition, system message.}
\small
\begin{quote}\ttfamily
You transcribe text from full-page document images.\\
Transcribe EXACTLY as written (accents, punctuation, casing, spacing).\\
Preserve line breaks and blank lines.\\
Do NOT correct spelling/grammar or normalize.\\
Do NOT describe the image or add preambles.\\
Reading order: multi-column pages are read column-by-column (left→right);\\
within a column, read top-to-bottom.\\
Tables: read row-by-row (top-to-bottom), cells left-to-right with a single
space between cells.\\
Output STRICT JSON only.
\end{quote}
\normalsize

\paragraph{User message.}
\small
\begin{quote}\ttfamily
Task: Transcribe the entire page at line level in the specified reading order.\\
Return ONLY valid JSON (no code fences, comments, or extra keys).\\
Output schema (JSON only): \\
\quad\{\,"lines": ["line\_001\_text", "line\_002\_text", "…"]\,\}
\end{quote}
\normalsize

\section{Additional Qualitative Examples}
\label{app:qual_examples}

\begin{figure*}[t]
  \centering

  \subfloat[\textbf{RIMES}]{
    \includegraphics[
      width=0.5\linewidth,
      height=0.5\textheight,
      keepaspectratio
    ]{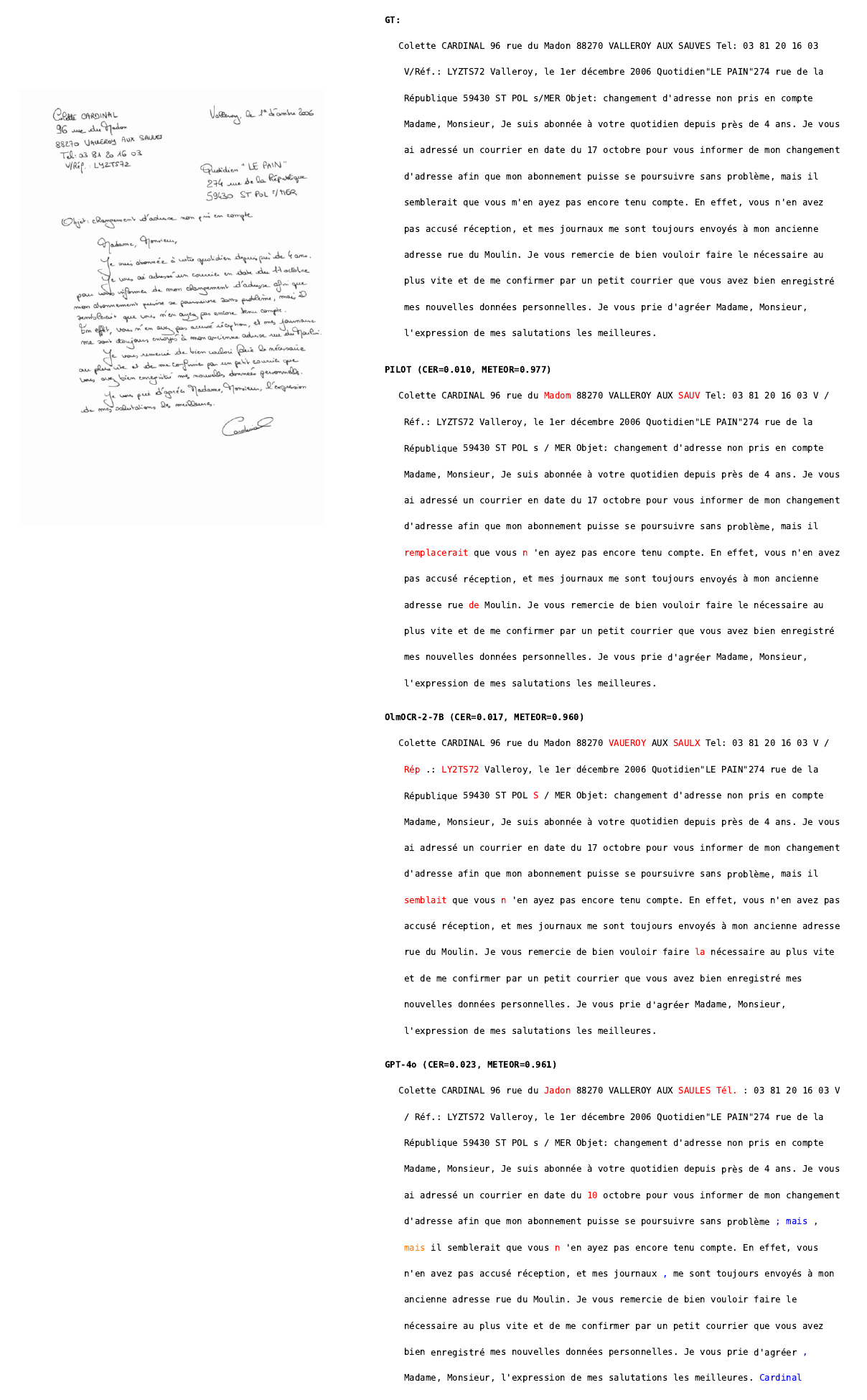}
    \label{fig:app_rimes}
  }
  \hfill
  \subfloat[\textbf{SROIE}]{
    \includegraphics[
      width=0.5\linewidth,
      height=0.5\textheight,
      keepaspectratio
    ]{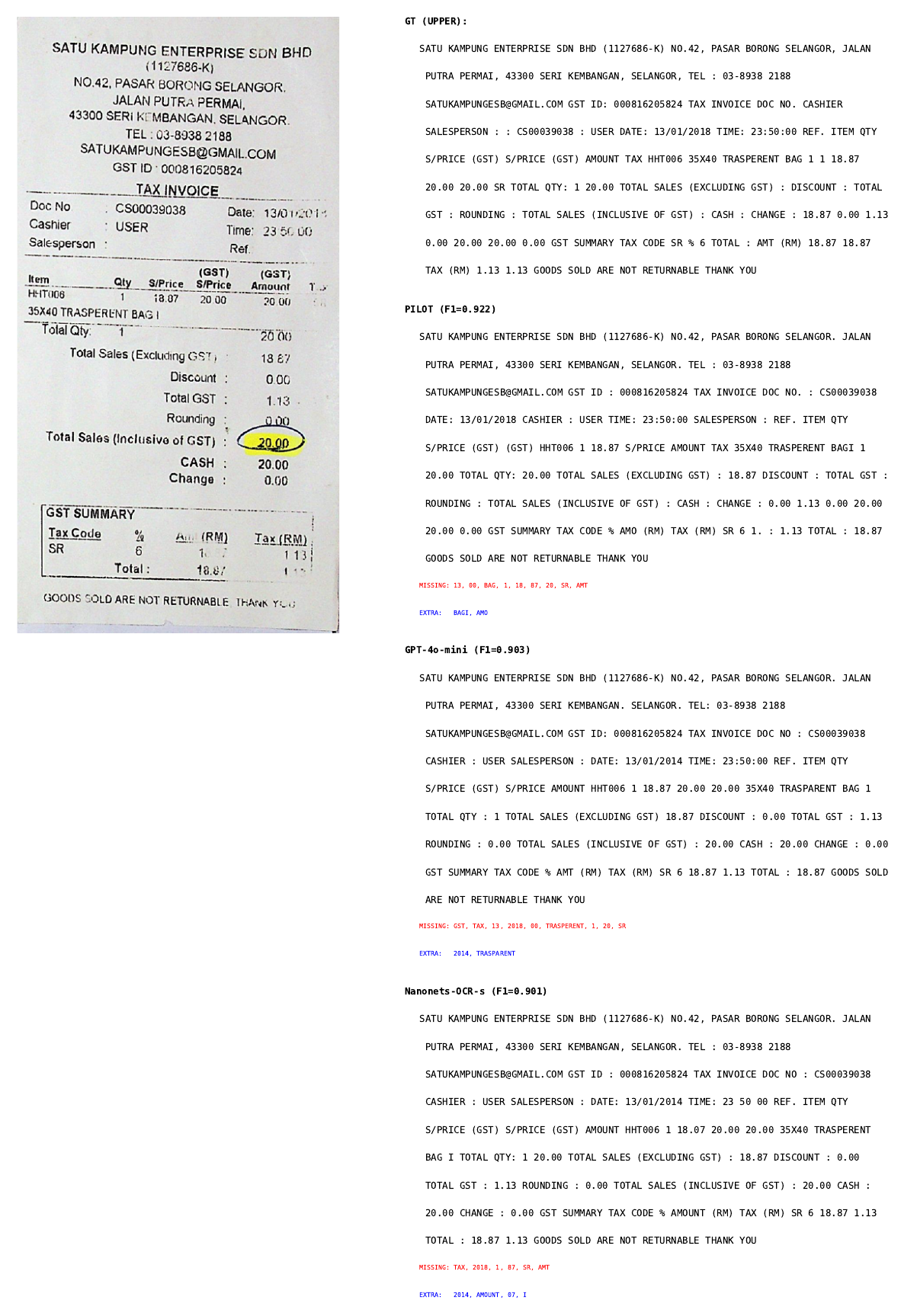}
    \label{fig:app_sroie}
  }

  \vspace{0.5em}

  \subfloat[\textbf{MAURDOR}]{
    \includegraphics[
      width=\linewidth,
      height=0.32\textheight,
      keepaspectratio
    ]{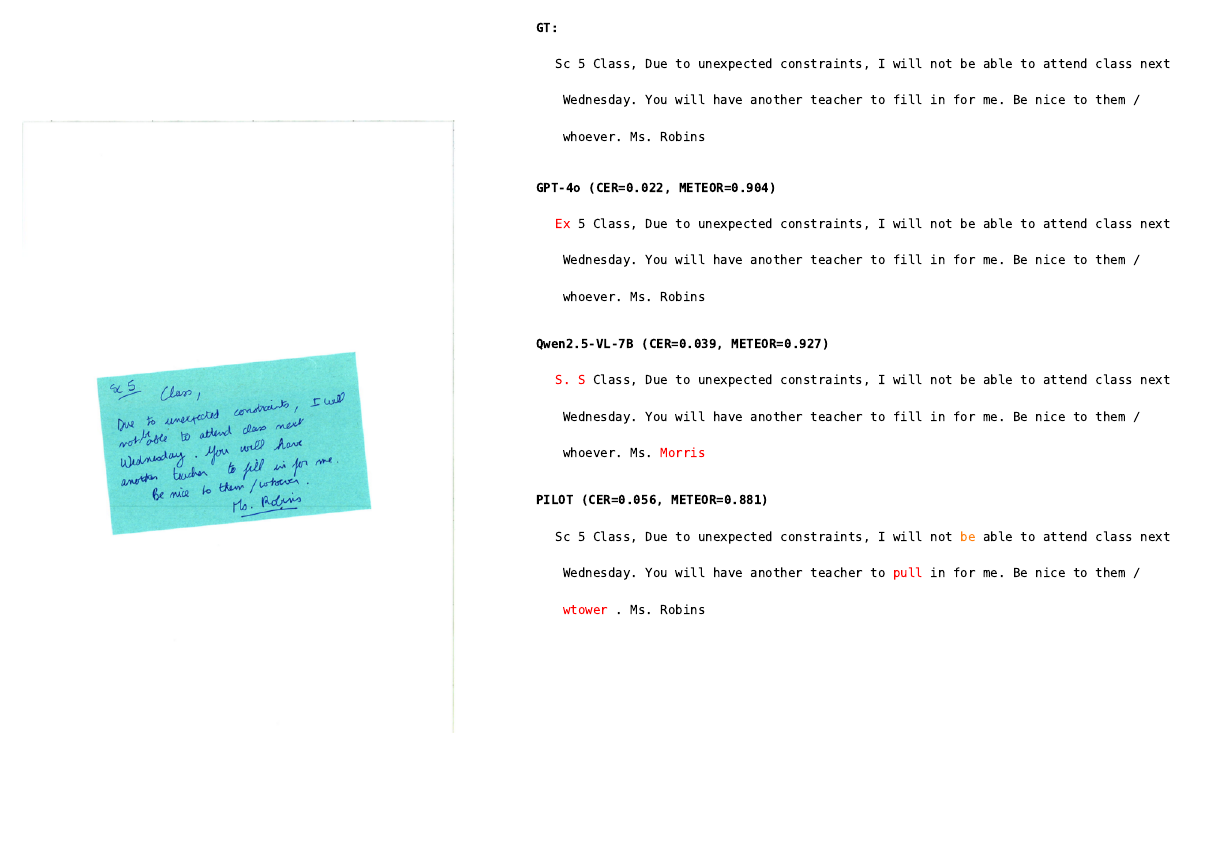}
    \label{fig:app_maurdor}
  }

  \caption{Qualitative OCR prediction samples on three datasets: RIMES, SROIE, and MAURDOR. For each example, the ground-truth text is shown together with model predictions and color-coded differences: black = correct token, red = substitution, blue = insertion, and orange = deletion.}
  \label{fig:app_qual_examples}
\end{figure*}

Qualitative OCR results. We show the ground-truth text at the top, followed by the top model predictions with color-coded differences: black = correct token, red = substitution, blue = insertion, and orange = deletion. See Figure~\ref{fig:app_qual_examples}.

\end{document}